# LatticeML: A data-driven application for predicting the effective Young Modulus of high temperature graph based architected materials

Akshansh Mishra[1,*]

[1]School of Industrial and Information Engineering, Politecnico Di Milano, Milan, Italy

Email: akshansh.mishra@mail.polimi.it

**Abstract:** Architected materials with their unique topology and geometry offer the potential to modify physical and mechanical properties. Machine learning can accelerate the design and optimization of these materials by identifying optimal designs and forecasting performance. This work presents LatticeML, a data-driven application for predicting the effective Young's Modulus of high-temperature graph-based architected materials. The study considers eleven graph-based lattice structures with two high-temperature alloys, Ti-6Al-4V and Inconel 625. Finite element simulations were used to compute the effective Young's Modulus of the 2x2x2 unit cell configurations. A machine learning framework was developed to predict the Young's Modulus, involving data collection, preprocessing, implementation of regression models, and deployment of the best-performing model. Five supervised learning algorithms were evaluated, with the XGBoost Regressor achieving the highest accuracy (MSE = 2.7993, MAE = 1.1521, R-squared = 0.9875). The application uses the Streamlit framework to create an interactive web interface, allowing users to input material and geometric parameters and obtain predicted Young's Modulus values.

**Keywords:** Architected Materials; Machine Learning; High temeperature alloys; Deployment

## 1. Introduction

Architected materials are a class of materials that modify their physical and mechanical properties through their unique topology and geometry. They draw inspiration from natural cellular materials like bones and corals, which possess special properties such as lightweight structures and topology-controlled mechanics [1-4]. Artificial architected materials, known as lattice structures, are designed to mimic these natural materials. The relative density is a crucial property, representing the ratio of the lattice volume to the bounding box volume. Architected materials with very low relative density (<5%) are considered foams, while those with 10-50% relative density are classified as lattice structures [5-7]. These materials can be classified based on their geometry as stochastic, periodic, or pseudo-periodic, with periodic lattice structures like the octet truss and Schwarz diamond being widely studied. The mechanical behavior of architected materials is either stretching-dominated, resulting in higher stiffness and strength, or bending-dominated, leading to lower stiffness but higher energy absorption [9-10]. This behavior is determined by the scaling law that relates their effective properties to relative density. Optimization methods such as functional gradation, hybrid designs, and higher-order lattices are used to enhance the mechanical performance of architected materials, providing the ability to tailor their relative density and geometry. Architected materials have applications in lightweight structures, energy absorption, thermal and acoustic insulation, biomedical implants, and other areas due to their unique physical and mechanical properties enabled by additive manufacturing.

Machine learning can significantly improve the design and optimization of architected materials using a variety of methods. It can help identify the best material designs and topologies to achieve desired properties like stiffness, strength, and energy absorption. By analyzing existing data on material properties and geometries, machine learning can forecast the performance of new designs and guide material selection and design choices. Machine learning can also help with the inverse design process, allowing designers to specify desired properties and generate designs that achieve them, which speeds up the discovery of new material architectures. Buehler et al. [11] proposed an unsupervised generative adversarial network (GAN) model for designing nature-inspired materials, which provides an alternative to traditional supervised deep learning methods. Without human intervention, the GAN model creates a latent space that can be explored and manipulated to create material designs. The method extends existing data distributions to create novel materials inspired by leaf microstructures in 2D and 3D dimensions. Peng et al. [12] proposed a data-efficient approach to optimizing the design of 3D-printed architected materials by incorporating a machine learning (ML) cycle that combines finite element method (FEM) simulations and 3D neural networks. This approach enables rational material design without the need for prior knowledge or extensive manual effort. The method was used in orthopedic implant design to create microscale heterogeneous architectures with a biocompatible elastic modulus and greater strength than uniform designs. Lee et al. [13] proposed a method for optimizing the elastic modulus, strength, and toughness of lattice structures while minimizing balances by focusing on the shape of beam elements. This method used generative deep learning to speed up the optimization process. The study emphasizes the importance of distributed stress fields and deformation modes in achieving significant improvements in mechanical properties depending on beam shape and lattice type. Wei et al. [14] developed a deep learning-based method for efficiently determining the elastic isotropy of architected materials directly from images of their unit cells, addressing the cost and time constraints of traditional methods. A measure of elastic isotropy for heterogeneous architected materials was created by building a database with corresponding unit cell images. Using connectivity, mechanical qualities, and stress levels of only 1% of nodes, Buehler et al. [15] employed a semi-supervised approach to construct topological structures of architected materials. In order to anticipate the distribution of load levels for the remaining 99% of nodes, this method made use of graph neural networks (GNNs), which learned graph embeddings and performed exceptionally well in semi-supervised classification tasks with little input.

The novelty of this work lies in the development of LatticeML, a data-driven machine learning application specifically designed to predict the effective Young's Modulus of high-temperature graph-based architected materials. Unlike previous studies that have employed machine learning for architected material design, this research focuses on a unique class of lattice structures and materials operating in extreme environments. The integration of finite element analysis and supervised learning algorithms within a user-friendly web application represents a significant advancement in the field. LatticeML enables rapid prediction of material properties, accelerating the design and optimization process without the need for extensive prior knowledge or manual effort.

## 2. Materials and Methods

In the present work, eleven graph-based lattice structures shown in Figure 1 were considered to have the material property of two high-temperature-based alloys i.e. Ti-6Al-4V and Inconel 625 whose material and thermal properties are shown in Table 1. The effective young modulus values of the 2 X 2 X 2 configuration of the given architected materials were found by carrying out the simulations on nTopology software. Six unit strain loads are utilized in the estimation of the unit cell's mechanical properties. Under periodic boundary conditions, these loads represent two separate forms of deformations: shear and tensile. The objective is to calculate the unit cell's effective stiffness matrix, C, which for an anisotropic material has 21 independent components. Three tensile loads (one for each axis) are applied in the X, Y, and Z directions. When a unit load is applied in the X direction, the block applies a strain of $\epsilon_{11}= 1$ while keeping $\epsilon_{22}=\epsilon_{33}=\epsilon_{12}=\epsilon_{13}=\epsilon_{23}= 0$ and similarly for Y and Z directions unit loads of $\epsilon_{22}= 1$ $and$ $\epsilon_{33}= 1$ are applied respectively. In the XY, XZ, and YZ planes (each plane separately), three shear loads are applied. The block applies a shear strain of $\gamma_{12} = 1$ while keeping the $\gamma_{13} = \gamma_{23} = 0$. Comparably, the unit loads applied to the XZ and YZ planes are $\gamma_{13} = 1$ and $\gamma_{23} = 1$. The system of equations that relates the stress tensor σ to the strain tensor ϵ through the anisotropic stiffness matrix C can be expressed as Equation 1.

$$\begin{bmatrix}\sigma_{11}\\ \sigma_{22}\\ \sigma_{33}\\ \sigma_{12}\\ \sigma_{13}\\ \sigma_{23}\end{bmatrix} = \begin{bmatrix}C_{11} & C_{12} & C_{13} & C_{14} & C_{15} & C_{16}\\ C_{21} & C_{22} & C_{23} & C_{24} & C_{25} & C_{26}\\ C_{31} & C_{32} & C_{33} & C_{34} & C_{35} & C_{36}\\ C_{41} & C_{42} & C_{43} & C_{44} & C_{45} & C_{46}\\ C_{51} & C_{52} & C_{53} & C_{54} & C_{55} & C_{56}\\ C_{61} & C_{62} & C_{63} & C_{64} & C_{65} & C_{66}\end{bmatrix}\begin{bmatrix}\epsilon_{11}\\ \epsilon_{22}\\ \epsilon_{33}\\ \gamma_{12}\\ \gamma_{13}\\ \gamma_{23}\end{bmatrix} \quad (1)$$

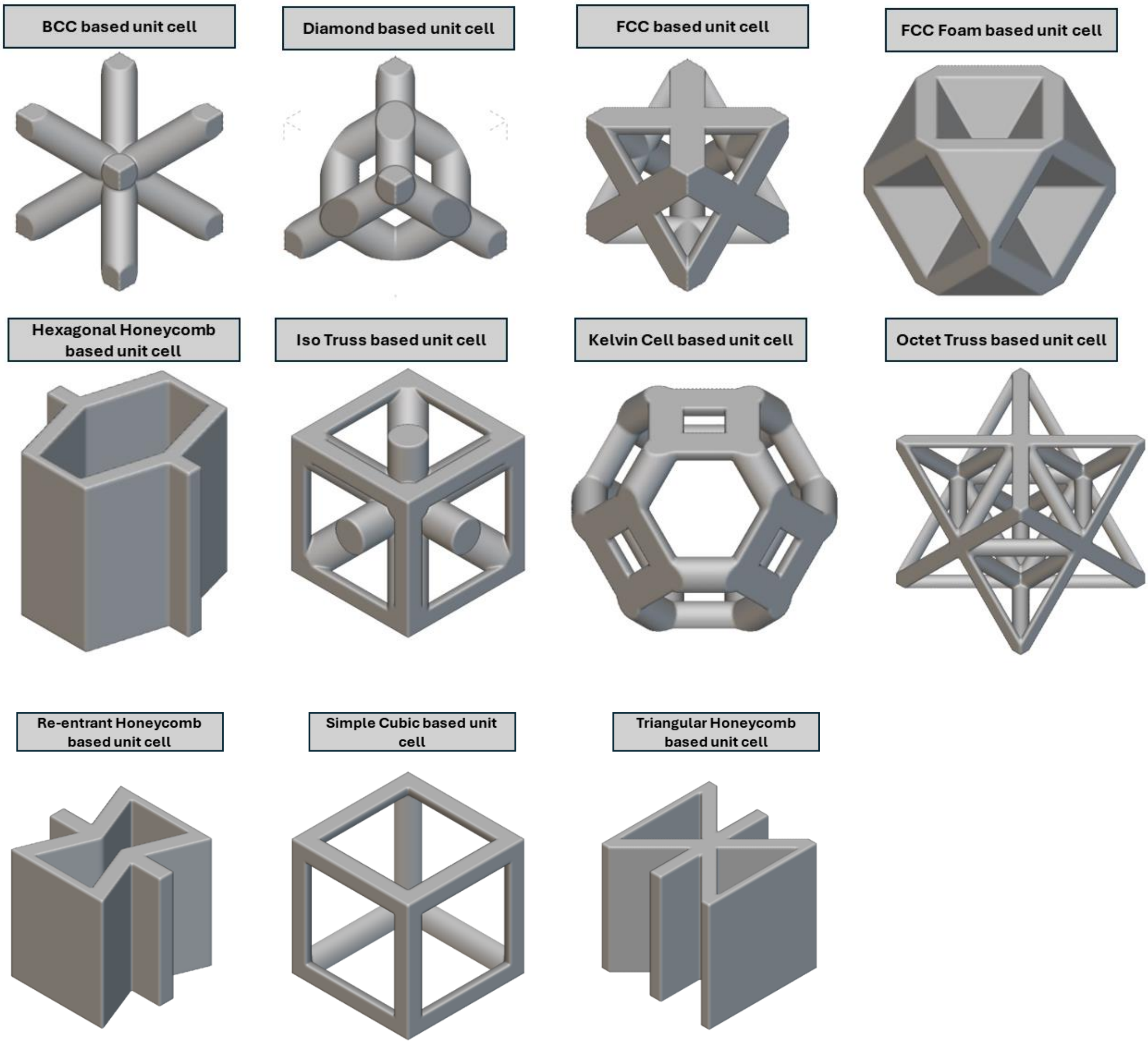

**Figure 1**. Graph based unit cells used in the present work

**Table 1.** Mechanical and thermal properties of the alloys considered in the present work

| Alloy | Elastic modulus (GPa) | Poisson ratio | Thermal conductivity (W/m.k) |
|---|---|---|---|
| Inconel 625 | 208 | 0.28 | 9.7 |
| Ti-6Al-4V | 138.8 | 0.34 | 6.7 |

Figure 2 shows the machine learning framework used in the present work. The flowchart depicts the process of using machine learning techniques to predict the effective Young's modulus of an architected material. The process begins with "Collecting the data from simulations," where data was collected for two types of high temperature-based alloys using

nTopology software. The next step is "Preparing the dataset," where the collected data was stored in a CSV file containing five input parameters and one output parameter having 110 datapoints shown in Table 2. This prepared dataset was then used in the subsequent step of “Implementing machine learning models.” Here, five supervised machine learning regression-based algorithms were employed to predict the effective Young’s modulus of the architected material. The flowchart then shows the “Finding the best model” step involved comparing the performance of the different models using metrics such as MSE, MAE, and R-square value, ultimately identifying the best-performing model.”Deployment of the best model,” which in this case was the LatticeML model. This model utilizes the Streamlit framework to create an interactive, data-driven web interface.

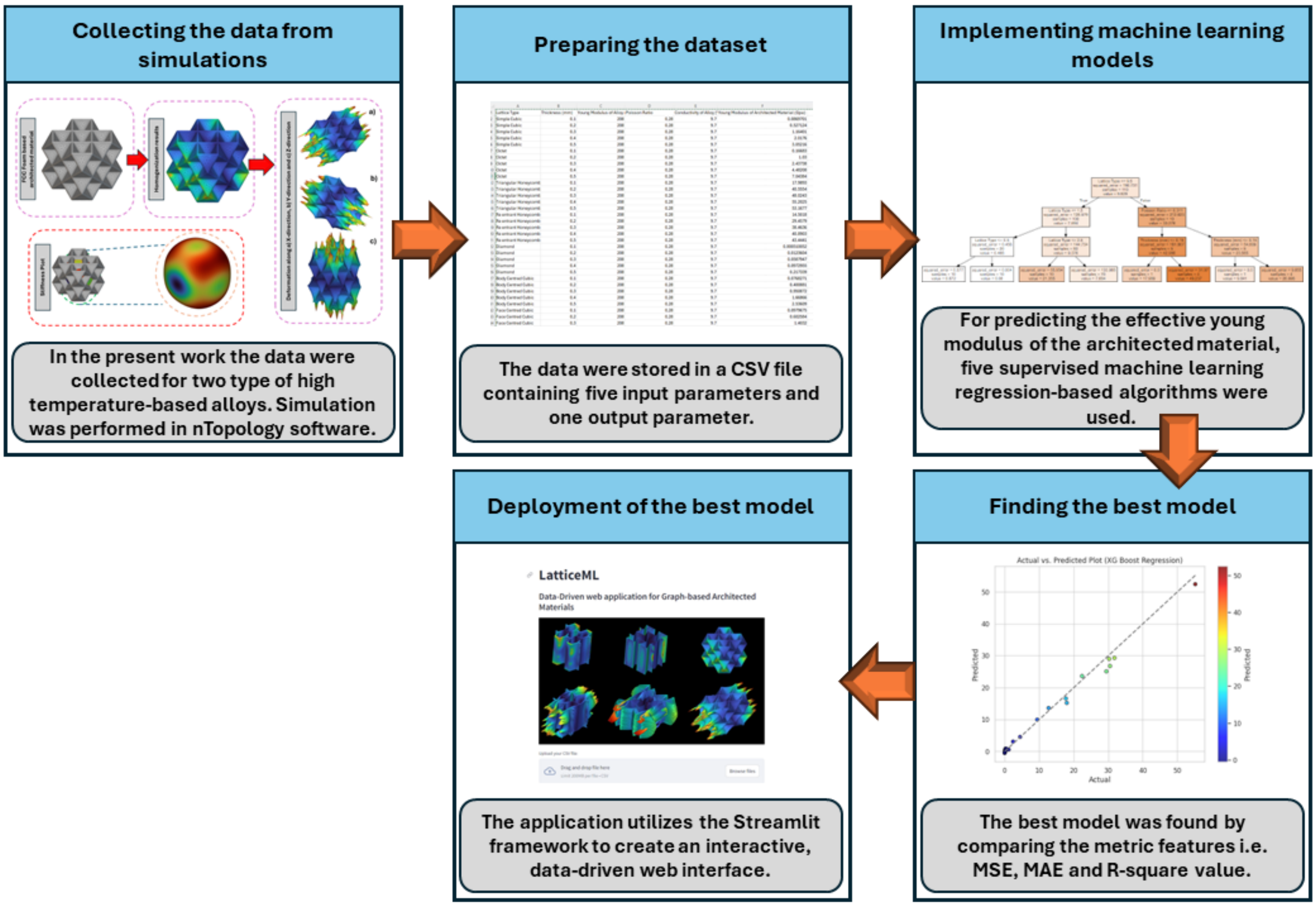

**Figure 2.** Machine learning framework used in the present work. The flowchart illustrates the step-by-step process of developing the LatticeML application, including data collection from finite element simulations, dataset preparation, implementation of regression models, identification of the best-performing model, and deployment of the final application using the Streamlit framework.

Streamlit is an open-source Python library that enables the creation of data applications and the deployment of machine learning models. The flow of data within Streamlit is a critical aspect, as illustrated in the accompanying Figure 3. When the source code is modified while a user is interacting with the application, two simultaneous processes occur in Streamlit. First, Streamlit executes all the callbacks present in the application. Afterwards, Streamlit runs the entire Python script from the top to the bottom, and the resulting output is then displayed in the user’s web browser.

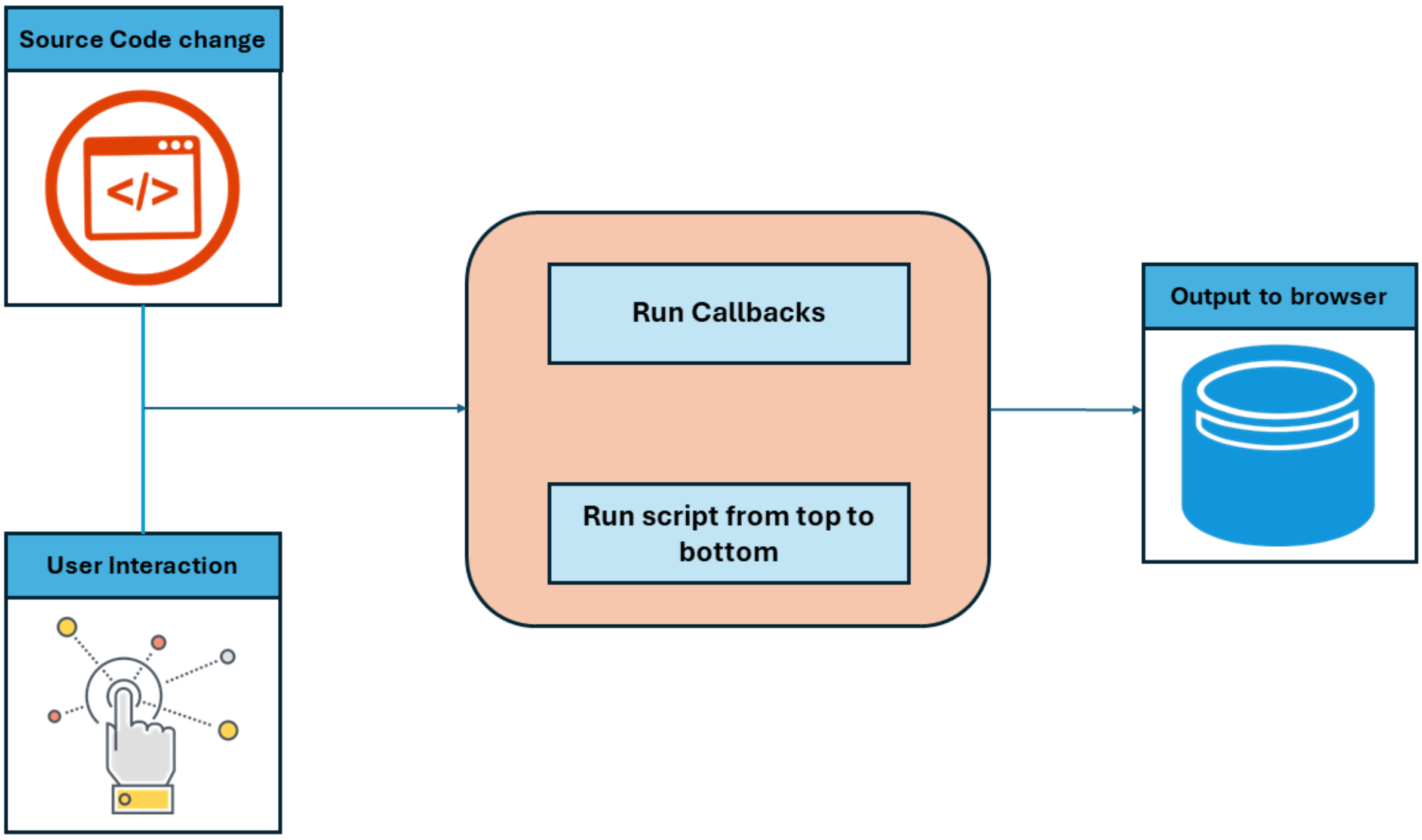

**Figure 3.** Data flow representation in Streamlit. This diagram depicts the key aspects of the Streamlit framework, highlighting how modifications to the source code trigger simultaneous execution of callbacks and the full Python script, resulting in the updated output being displayed in the user's web browser.

**Table 2.** Obtained data from the simulations

| Lattice Type | Thickness (mm) | Young Modulus of Alloy (GPa) | Poisson Ratio | Conductivity of Alloy (W/m.K) | Young Modulus of Architected Material (GPa) |
|---|---|---|---|---|---|
| Simple Cubic | 0.1 | 208 | 0.28 | 9.7 | 0.0869701 |
| Simple Cubic | 0.2 | 208 | 0.28 | 9.7 | 0.527124 |
| Simple Cubic | 0.3 | 208 | 0.28 | 9.7 | 1.16401 |
| Simple Cubic | 0.4 | 208 | 0.28 | 9.7 | 2.0176 |
| Simple Cubic | 0.5 | 208 | 0.28 | 9.7 | 3.05216 |
| Octet | 0.1 | 208 | 0.28 | 9.7 | 0.16683 |
| Octet | 0.2 | 208 | 0.28 | 9.7 | 1.03 |
| Octet | 0.3 | 208 | 0.28 | 9.7 | 2.43738 |
| Octet | 0.4 | 208 | 0.28 | 9.7 | 4.48208 |
| Octet | 0.5 | 208 | 0.28 | 9.7 | 7.04384 |
| Triangular Honeycomb | 0.1 | 208 | 0.28 | 9.7 | 17.9893 |
| Triangular Honeycomb | 0.2 | 208 | 0.28 | 9.7 | 40.5554 |
| Triangular Honeycomb | 0.3 | 208 | 0.28 | 9.7 | 48.0243 |
| Triangular Honeycomb | 0.4 | 208 | 0.28 | 9.7 | 55.2025 |
| Triangular Honeycomb | 0.5 | 208 | 0.28 | 9.7 | 53.1677 |
| Re entrant Honeycomb | 0.1 | 208 | 0.28 | 9.7 | 14.5018 |
| Re entrant Honeycomb | 0.2 | 208 | 0.28 | 9.7 | 29.4579 |
| Re entrant Honeycomb | 0.3 | 208 | 0.28 | 9.7 | 38.4636 |
| Re entrant Honeycomb | 0.4 | 208 | 0.28 | 9.7 | 40.8903 |
| Re entrant Honeycomb | 0.5 | 208 | 0.28 | 9.7 | 43.4441 |
| Diamond | 0.1 | 208 | 0.28 | 9.7 | 0.000510052 |
| Diamond | 0.2 | 208 | 0.28 | 9.7 | 0.0123604 |
| Diamond | 0.3 | 208 | 0.28 | 9.7 | 0.0587947 |
| Diamond | 0.4 | 208 | 0.28 | 9.7 | 0.0972955 |
| Diamond | 0.5 | 208 | 0.28 | 9.7 | 0.217339 |
| Body Centred Cubic | 0.1 | 208 | 0.28 | 9.7 | 0.0768271 |
| Body Centred Cubic | 0.2 | 208 | 0.28 | 9.7 | 0.400881 |
| Body Centred Cubic | 0.3 | 208 | 0.28 | 9.7 | 0.950872 |
| Body Centred Cubic | 0.4 | 208 | 0.28 | 9.7 | 1.66866 |
| Body Centred Cubic | 0.5 | 208 | 0.28 | 9.7 | 2.53609 |
| Face Centred Cubic | 0.1 | 208 | 0.28 | 9.7 | 0.0979675 |
| Face Centred Cubic | 0.2 | 208 | 0.28 | 9.7 | 0.602584 |
| Face Centred Cubic | 0.3 | 208 | 0.28 | 9.7 | 1.4032 |
| Face Centred Cubic | 0.4 | 208 | 0.28 | 9.7 | 2.54084 |
| Face Centred Cubic | 0.5 | 208 | 0.28 | 9.7 | 3.88883 |
| Hexagonal Honeycomb | 0.1 | 208 | 0.28 | 9.7 | 11.658 |
| Hexagonal Honeycomb | 0.2 | 208 | 0.28 | 9.7 | 23.2304 |
| Hexagonal Honeycomb | 0.3 | 208 | 0.28 | 9.7 | 29.8872 |
| Hexagonal Honeycomb | 0.4 | 208 | 0.28 | 9.7 | 35.7401 |
| Hexagonal Honeycomb | 0.5 | 208 | 0.28 | 9.7 | 35.2737 |
| Kelvin Cell | 0.1 | 208 | 0.28 | 9.7 | 0.000531771 |
| Kelvin Cell | 0.2 | 208 | 0.28 | 9.7 | 0.0165469 |
| Kelvin Cell | 0.3 | 208 | 0.28 | 9.7 | 0.092296 |
| Kelvin Cell | 0.4 | 208 | 0.28 | 9.7 | 0.307433 |
| Kelvin Cell | 0.5 | 208 | 0.28 | 9.7 | 0.705788 |
| Iso Truss | 0.1 | 208 | 0.28 | 9.7 | 0.151367 |
| Iso Truss | 0.2 | 208 | 0.28 | 9.7 | 0.901676 |
| Iso Truss | 0.3 | 208 | 0.28 | 9.7 | 2.07243 |
| Iso Truss | 0.4 | 208 | 0.28 | 9.7 | 3.68043 |
| Iso Truss | 0.5 | 208 | 0.28 | 9.7 | 5.50209 |
| FCC Foam | 0.1 | 208 | 0.28 | 9.7 | 16.9701 |
| FCC Foam | 0.2 | 208 | 0.28 | 9.7 | 28.2643 |
| FCC Foam | 0.3 | 208 | 0.28 | 9.7 | 30.5209 |
| FCC Foam | 0.4 | 208 | 0.28 | 9.7 | 29.6728 |
| FCC Foam | 0.5 | 208 | 0.28 | 9.7 | 31.8022 |
| Simple Cubic | 0.1 | 138.8 | 0.342 | 6.7 | 0.047599 |
| Simple Cubic | 0.2 | 138.8 | 0.342 | 6.7 | 0.288788 |
| Simple Cubic | 0.3 | 138.8 | 0.342 | 6.7 | 0.637965 |
| Simple Cubic | 0.4 | 138.8 | 0.342 | 6.7 | 1.10666 |
| Simple Cubic | 0.5 | 138.8 | 0.342 | 6.7 | 1.67495 |
| Octet | 0.1 | 138.8 | 0.342 | 6.7 | 0.0913379 |
| Octet | 0.2 | 138.8 | 0.342 | 6.7 | 0.564378 |
| Octet | 0.3 | 138.8 | 0.342 | 6.7 | 1.33652 |
| Octet | 0.4 | 138.8 | 0.342 | 6.7 | 2.45951 |
| Octet | 0.5 | 138.8 | 0.342 | 6.7 | 3.86823 |
| Triangular Honeycomb | 0.1 | 138.8 | 0.342 | 6.7 | 9.84095 |
| Triangular Honeycomb | 0.2 | 138.8 | 0.342 | 6.7 | 22.2262 |
| Triangular Honeycomb | 0.3 | 138.8 | 0.342 | 6.7 | 26.3193 |
| Triangular Honeycomb | 0.4 | 138.8 | 0.342 | 6.7 | 30.2602 |
| Triangular Honeycomb | 0.5 | 138.8 | 0.342 | 6.7 | 29.1769 |
| Re entrant Honeycomb | 0.1 | 138.8 | 0.342 | 6.7 | 7.93576 |
| Re entrant Honeycomb | 0.2 | 138.8 | 0.342 | 6.7 | 16.1398 |
| Re entrant Honeycomb | 0.3 | 138.8 | 0.342 | 6.7 | 21.0905 |
| Re entrant Honeycomb | 0.4 | 138.8 | 0.342 | 6.7 | 22.4164 |
| Re entrant Honeycomb | 0.5 | 138.8 | 0.342 | 6.7 | 23.8383 |
| Diamond | 0.1 | 138.8 | 0.342 | 6.7 | 0.000279763 |
| Diamond | 0.2 | 138.8 | 0.342 | 6.7 | 0.00679377 |
| Diamond | 0.3 | 138.8 | 0.342 | 6.7 | 0.0323304 |
| Diamond | 0.4 | 138.8 | 0.342 | 6.7 | 0.0533401 |
| Diamond | 0.5 | 138.8 | 0.342 | 6.7 | 0.11957 |
| Body Centred Cubic | 0.1 | 138.8 | 0.342 | 6.7 | 0.0420683 |
| Body Centred Cubic | 0.2 | 138.8 | 0.342 | 6.7 | 0.219526 |
| Body Centred Cubic | 0.3 | 138.8 | 0.342 | 6.7 | 0.520995 |
| Body Centred Cubic | 0.4 | 138.8 | 0.342 | 6.7 | 0.914647 |
| Body Centred Cubic | 0.5 | 138.8 | 0.342 | 6.7 | 1.39051 |
| Face Centred Cubic | 0.1 | 138.8 | 0.342 | 6.7 | 0.0536066 |
| Face Centred Cubic | 0.2 | 138.8 | 0.342 | 6.7 | 0.32969 |
| Face Centred Cubic | 0.3 | 138.8 | 0.342 | 6.7 | 0.767773 |
| Face Centred Cubic | 0.4 | 138.8 | 0.342 | 6.7 | 1.39021 |
| Face Centred Cubic | 0.5 | 138.8 | 0.342 | 6.7 | 2.12855 |
| Hexagonal Honeycomb | 0.1 | 138.8 | 0.342 | 6.7 | 6.3807 |
| Hexagonal Honeycomb | 0.2 | 138.8 | 0.342 | 6.7 | 12.7312 |
| Hexagonal Honeycomb | 0.3 | 138.8 | 0.342 | 6.7 | 16.3865 |
| Hexagonal Honeycomb | 0.4 | 138.8 | 0.342 | 6.7 | 19.6004 |
| Hexagonal Honeycomb | 0.5 | 138.8 | 0.342 | 6.7 | 19.3486 |
| Kelvin Cell | 0.1 | 138.8 | 0.342 | 6.7 | 0.000291577 |
| Kelvin Cell | 0.2 | 138.8 | 0.342 | 6.7 | 0.0090961 |
| Kelvin Cell | 0.3 | 138.8 | 0.342 | 6.7 | 0.0508268 |
| Kelvin Cell | 0.4 | 138.8 | 0.342 | 6.7 | 0.169576 |
| Kelvin Cell | 0.5 | 138.8 | 0.342 | 6.7 | 0.389164 |
| Iso Truss | 0.1 | 138.8 | 0.342 | 6.7 | 0.0828692 |
| Iso Truss | 0.2 | 138.8 | 0.342 | 6.7 | 0.494307 |
| Iso Truss | 0.3 | 138.8 | 0.342 | 6.7 | 1.13721 |
| Iso Truss | 0.4 | 138.8 | 0.342 | 6.7 | 2.02136 |
| Iso Truss | 0.5 | 138.8 | 0.342 | 6.7 | 3.02333 |
| FCC Foam | 0.1 | 138.8 | 0.342 | 6.7 | 9.42718 |
| FCC Foam | 0.2 | 138.8 | 0.342 | 6.7 | 15.7062 |
| FCC Foam | 0.3 | 138.8 | 0.342 | 6.7 | 16.9339 |
| FCC Foam | 0.4 | 138.8 | 0.342 | 6.7 | 16.5215 |
| FCC Foam | 0.5 | 138.8 | 0.342 | 6.7 | 17.7296 |

## 3. Results and Discussion

### 3.1. Collection of the data from simulations

In mechanics, homogenization is an effective mathematical and computational method that is especially useful when studying composite materials. The goal is to understand the effective behavior of a material consisting of several, frequently heterogeneous, constituents on a macroscopic level. At the microscopic scale, the spatial domain of microstructure is represented by $\Omega_{\varepsilon}$ , where $\varepsilon$ represents the characteristics length scale associated with the microstructure. At this scale, the presence of distinct constituents, defects, or geometric features can cause the material properties to vary significantly from one point to another. The governing equations for the behavior of the material at the microscopic level can be expressed as shown in Equation 2. The equilibrium of forces within the material's microstructure is expressed by this equation.

$$\nabla.\sigma^{\varepsilon} + f = 0 \quad (2)$$

Where $\sigma^{\varepsilon}$ represents the stress tensor and $f$ represents the external forces or sources.

Using asymptotic methods, these microscopic equations are expanded, and the solution is expressed as a series in terms of $\varepsilon$ as shown in Equation 3.

$$u^{\varepsilon}(x) = u^{0}(x) + \varepsilon u^{1}(x) + \varepsilon^{2} u^{2}(x) + \cdots \quad (3)$$

Where $u^{\varepsilon}(x)$ represents the displacement field at the microscale.

The equations governing the macroscale behavior are then derived using the method of multiple scales, which involves averaging over the periodic microstructure. Equations that characterize the behavior of the material at a larger scale are homogenized as a result of this process. Predicting macroscopic behavior from microstructural properties, the homogenized equations usually take the form of partial differential equations relating effective stress and strain quantities as shown in Equation 4.

$$\nabla.\sigma + F = 0 \quad (4)$$

Where $\sigma$ is homogenized stress tensor, and $F$ represents effective external forces or sources. Figure 4 shows the obtained results from the homogenization.

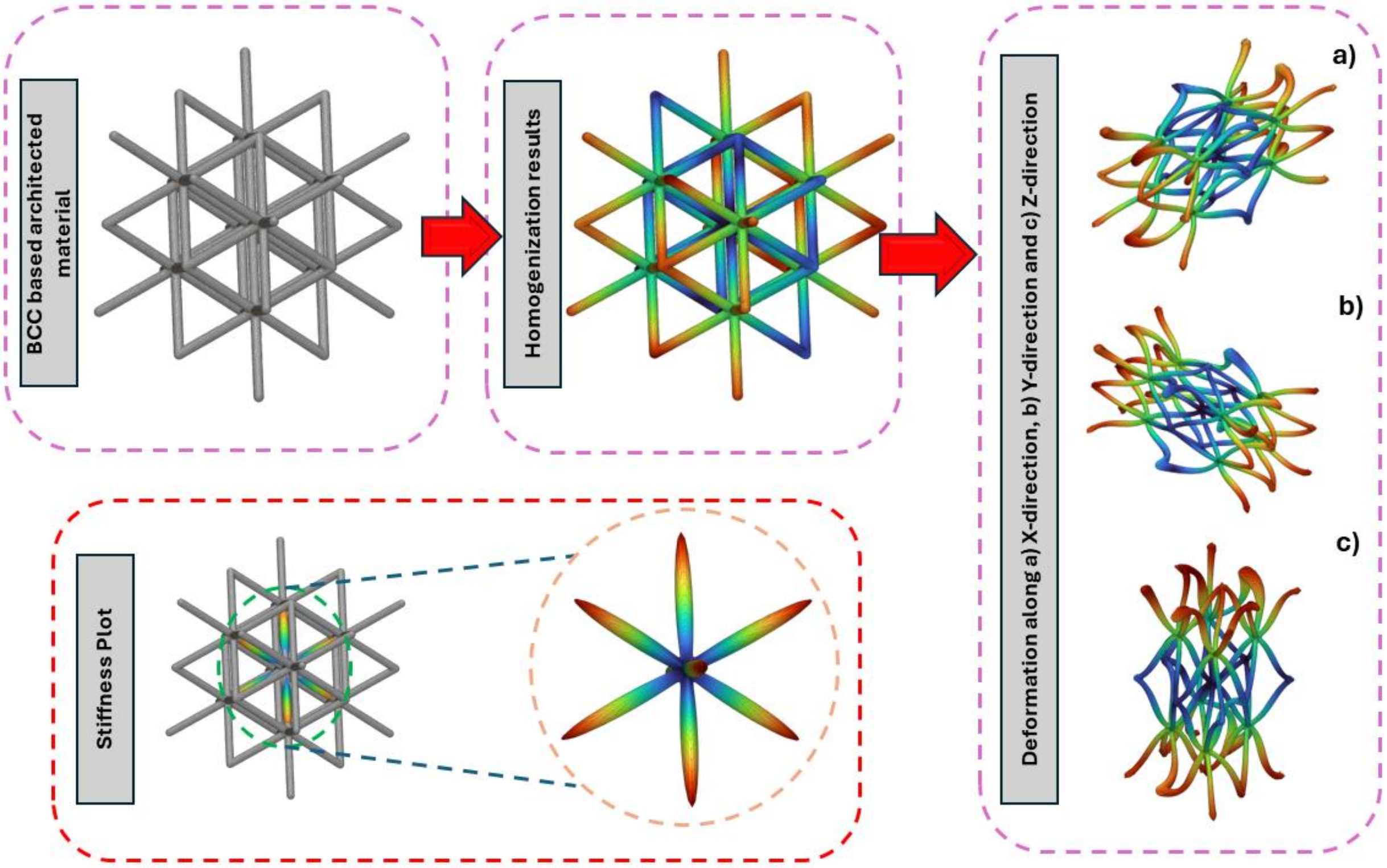

a)

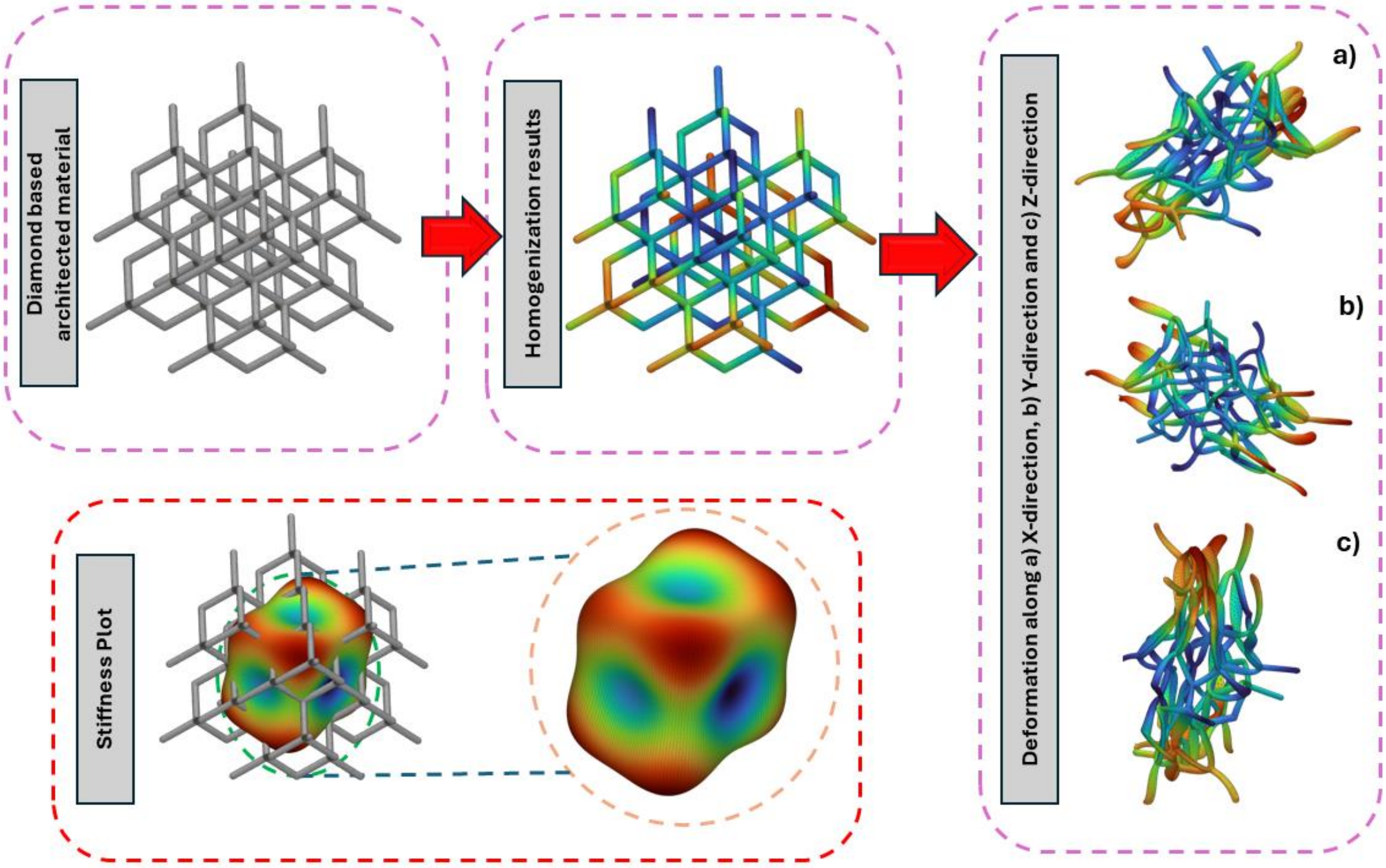
Diamond based architected material
Homogenization results
Deformation along a) X-direction, b) Y-direction and c) Z-direction
a)
b)
c)
Stiffness Plot

b)

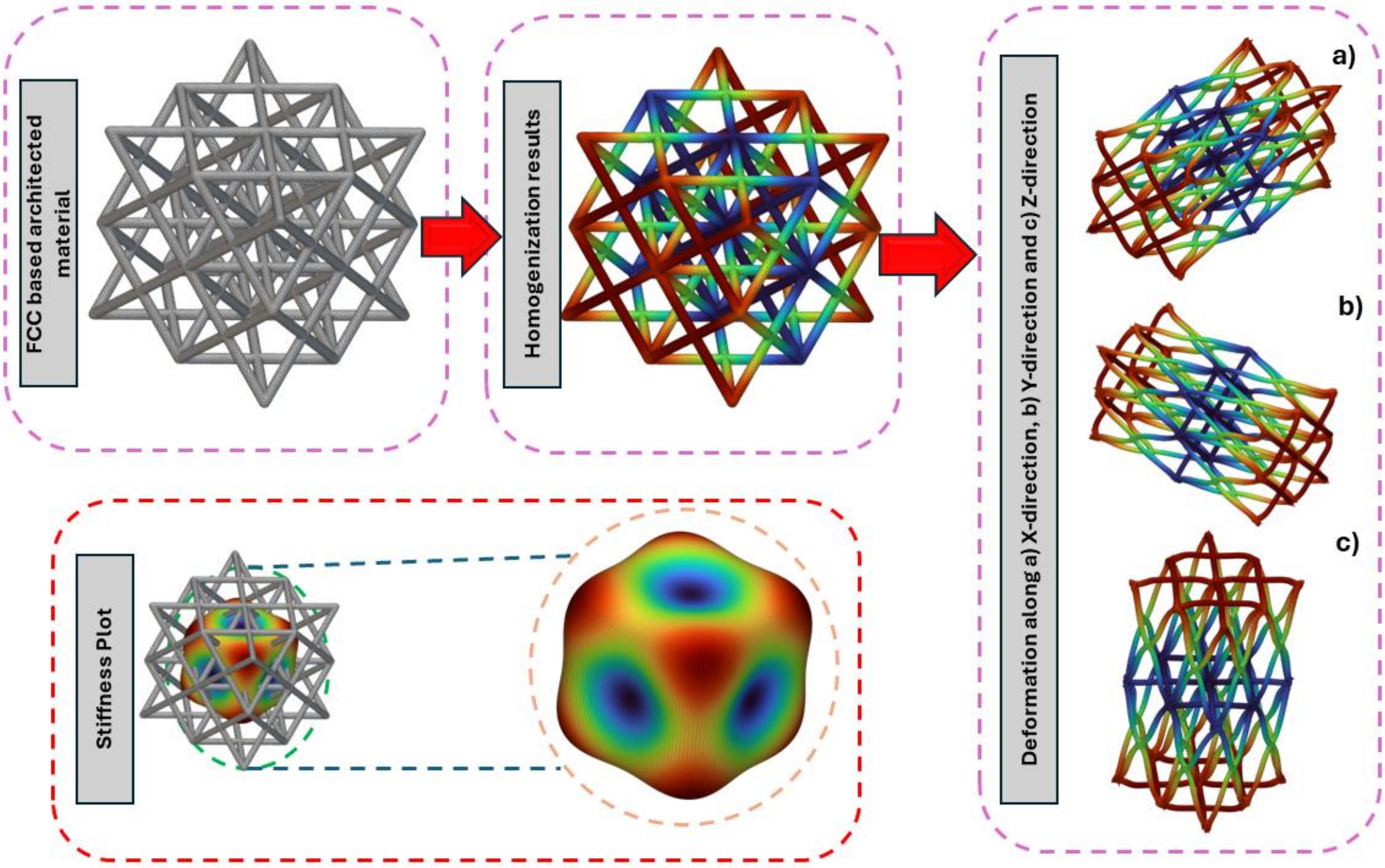
FCC based architected material
Homogenization results
Deformation along a) X-direction, b) Y-direction and c) Z-direction
a)
b)
c)
Stiffness Plot

c)

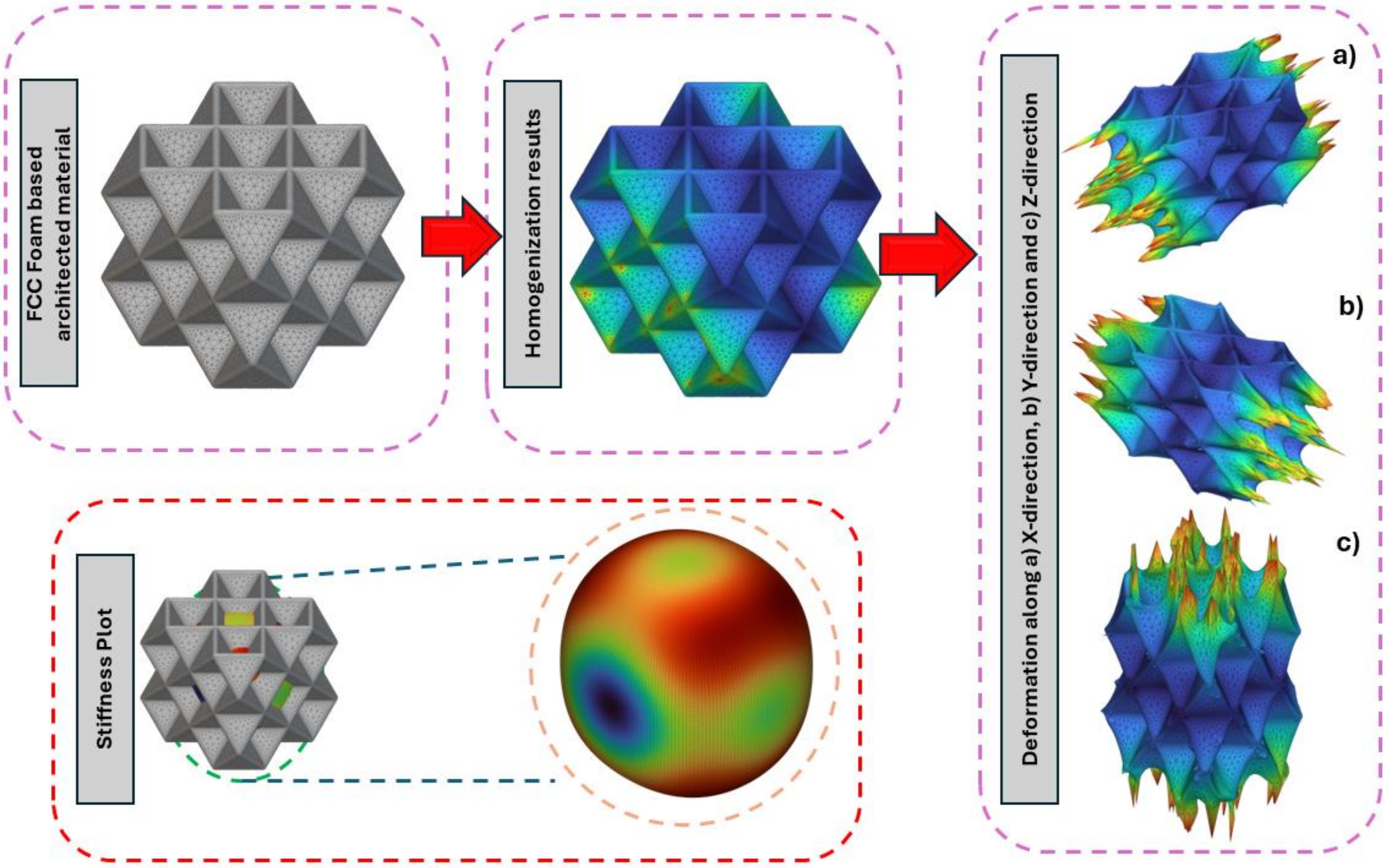
FCC Foam based architected material
Homogenization results
Deformation along a) X-direction, b) Y-direction and c) Z-direction
a)
b)
c)
Stiffness Plot

d)

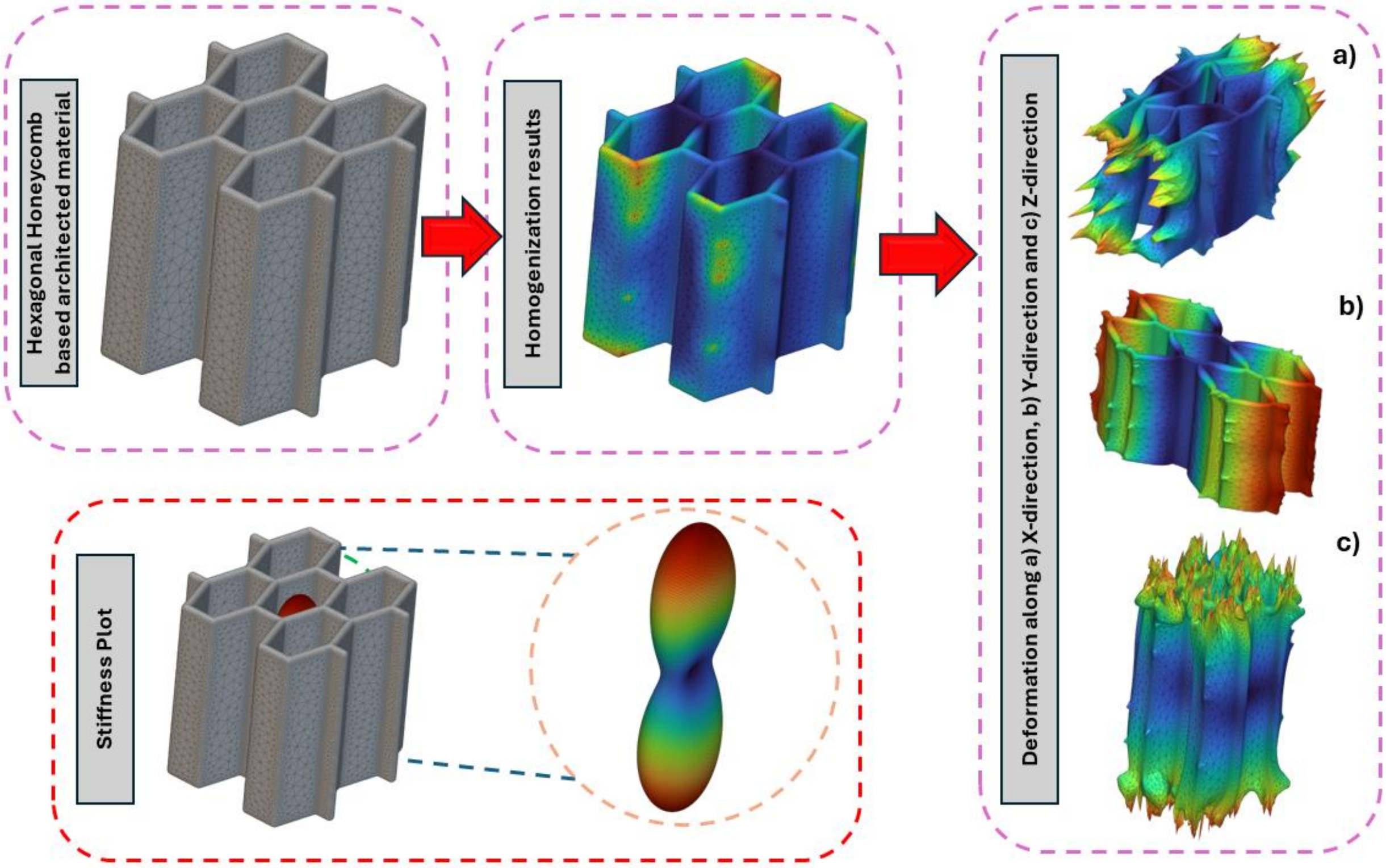
Hexagonal Honeycomb based architected material
Homogenization results
Deformation along a) X-direction, b) Y-direction and c) Z-direction
a)
b)
c)
Stiffness Plot

e)

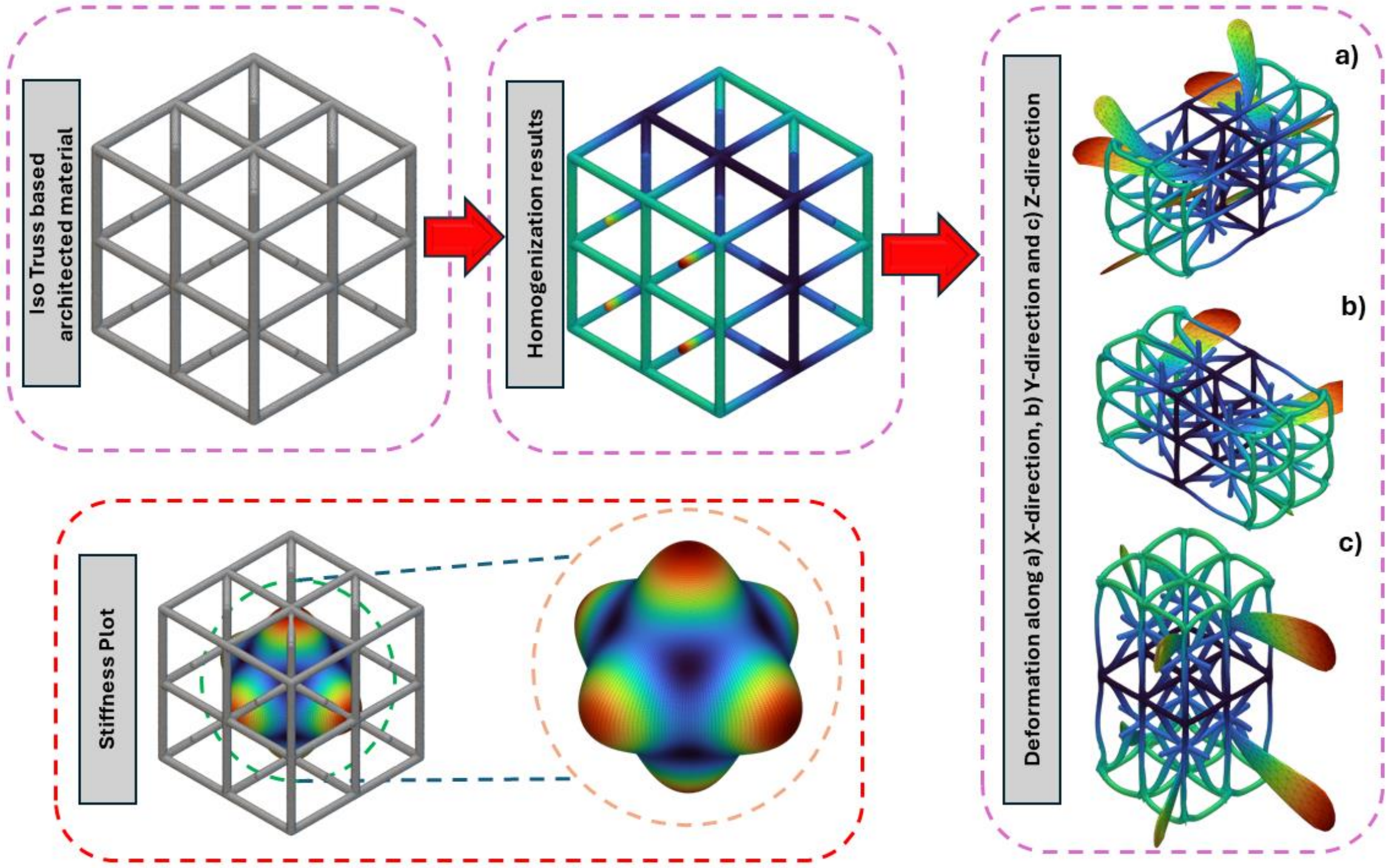
Iso Truss based architected material
Homogenization results
Deformation along a) X-direction, b) Y-direction and c) Z-direction
a)
b)
c)
Stiffness Plot

f)

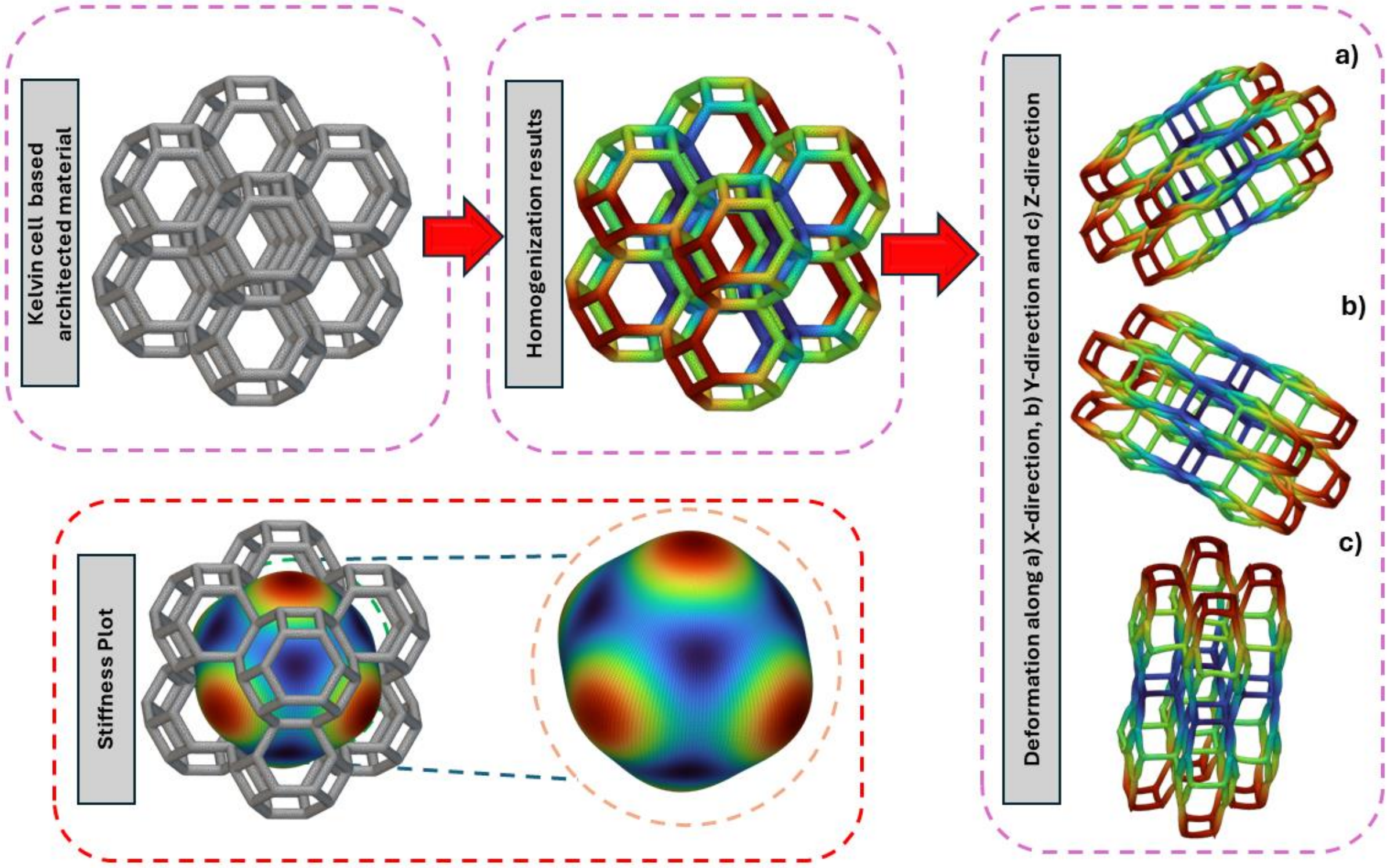
Kelvin cell based architected material
Homogenization results
Deformation along a) X-direction, b) Y-direction and c) Z-direction
a)
b)
c)
Stiffness Plot

g)

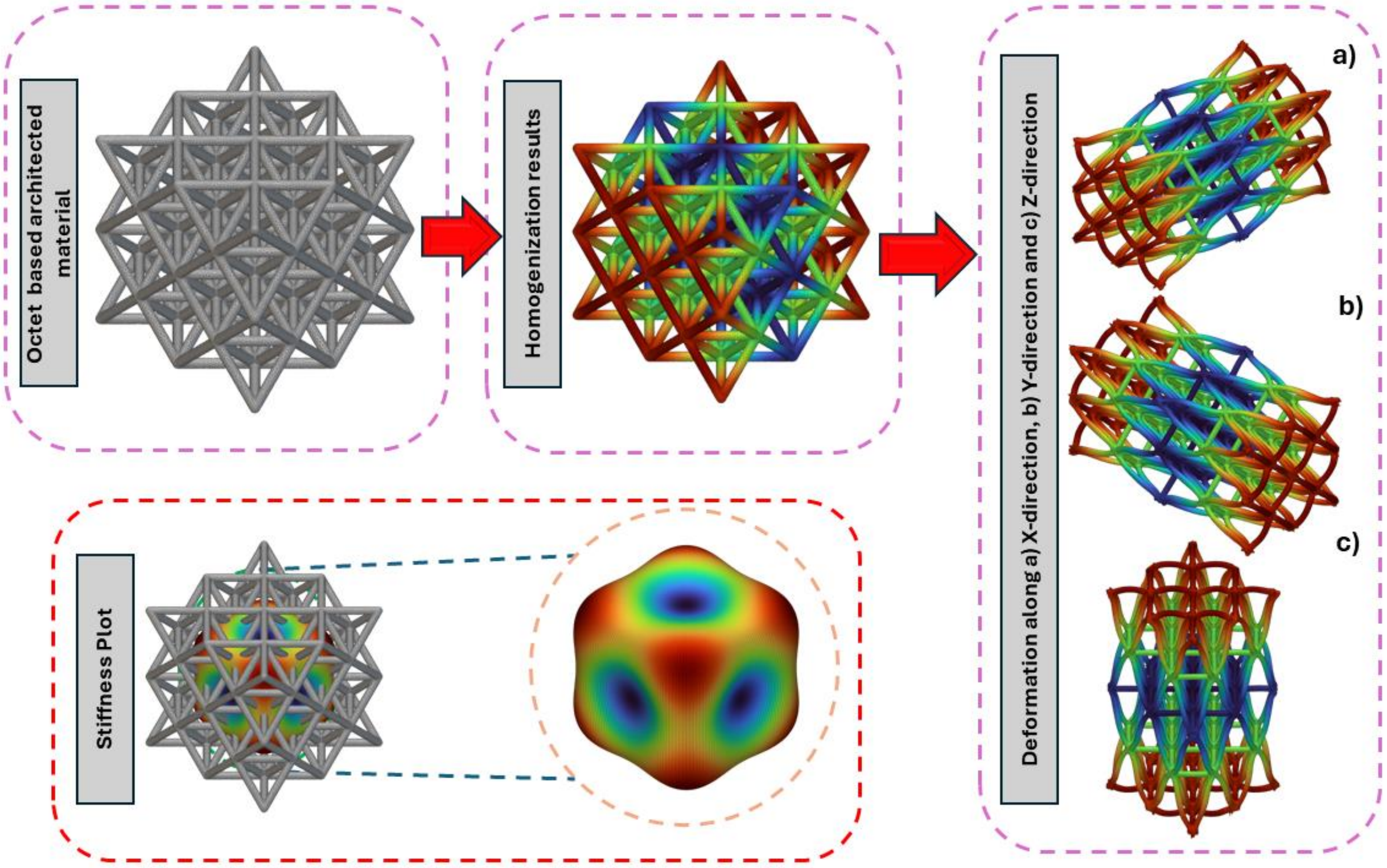
Octet based architected material
Homogenization results
Deformation along a) X-direction, b) Y-direction and c) Z-direction
a)
b)
c)
Stiffness Plot

h)

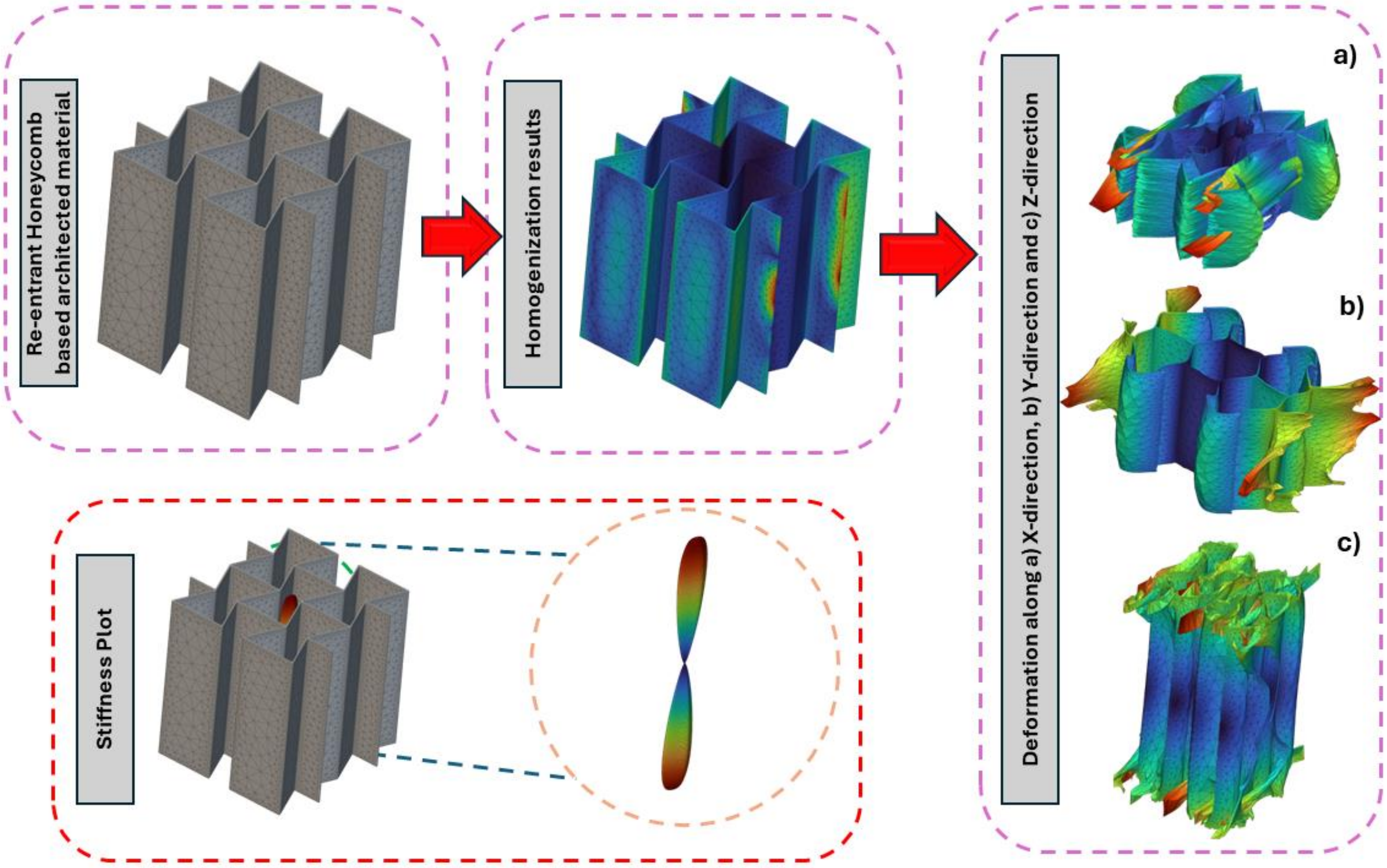
Re-entrant Honeycomb based architected material
Homogenization results
Deformation along a) X-direction, b) Y-direction and c) Z-direction
a)
b)
c)
Stiffness Plot

i)

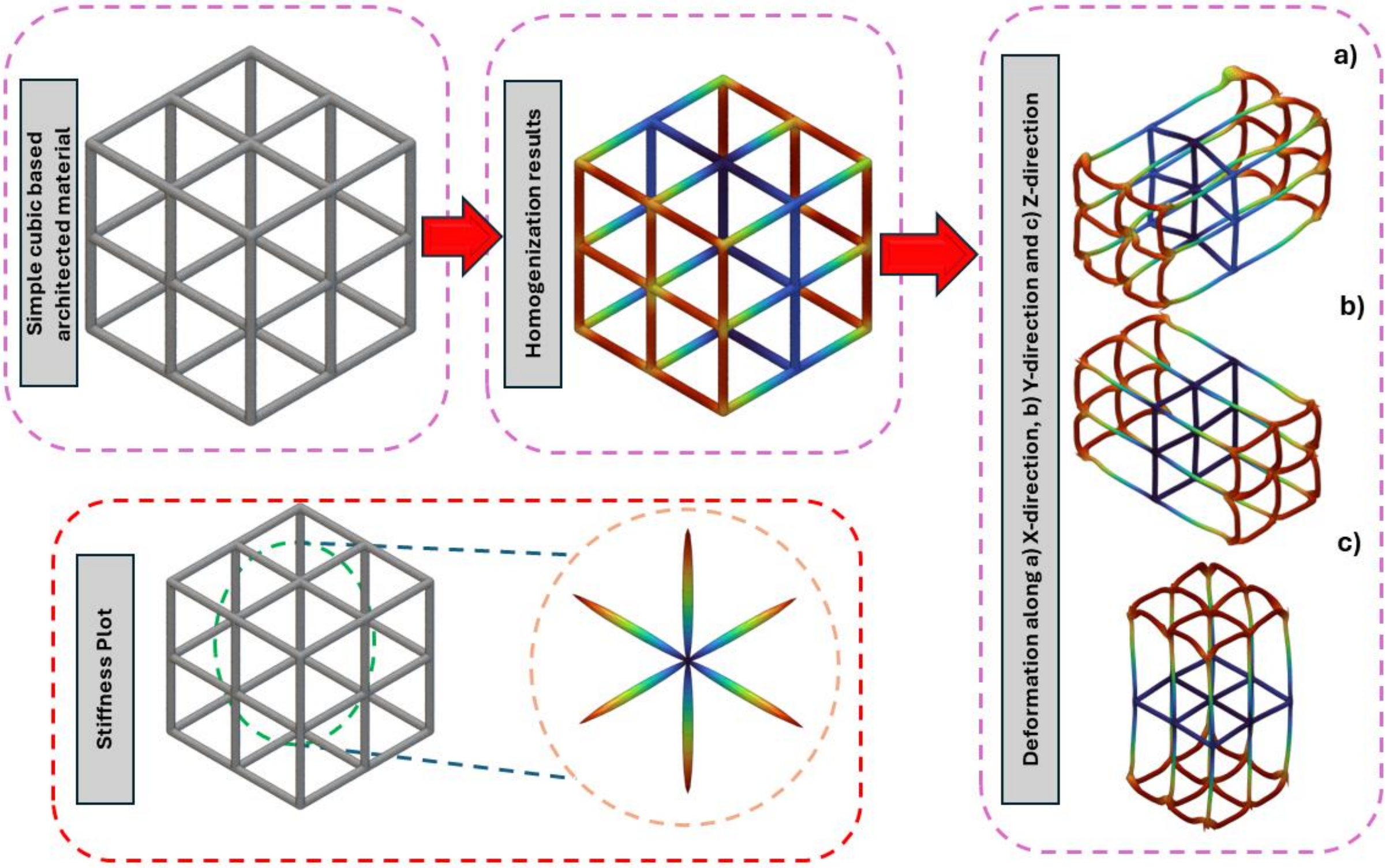
Simple cubic based architected material
Homogenization results
Deformation along a) X-direction, b) Y-direction and c) Z-direction
a)
b)
c)
Stiffness Plot

j)

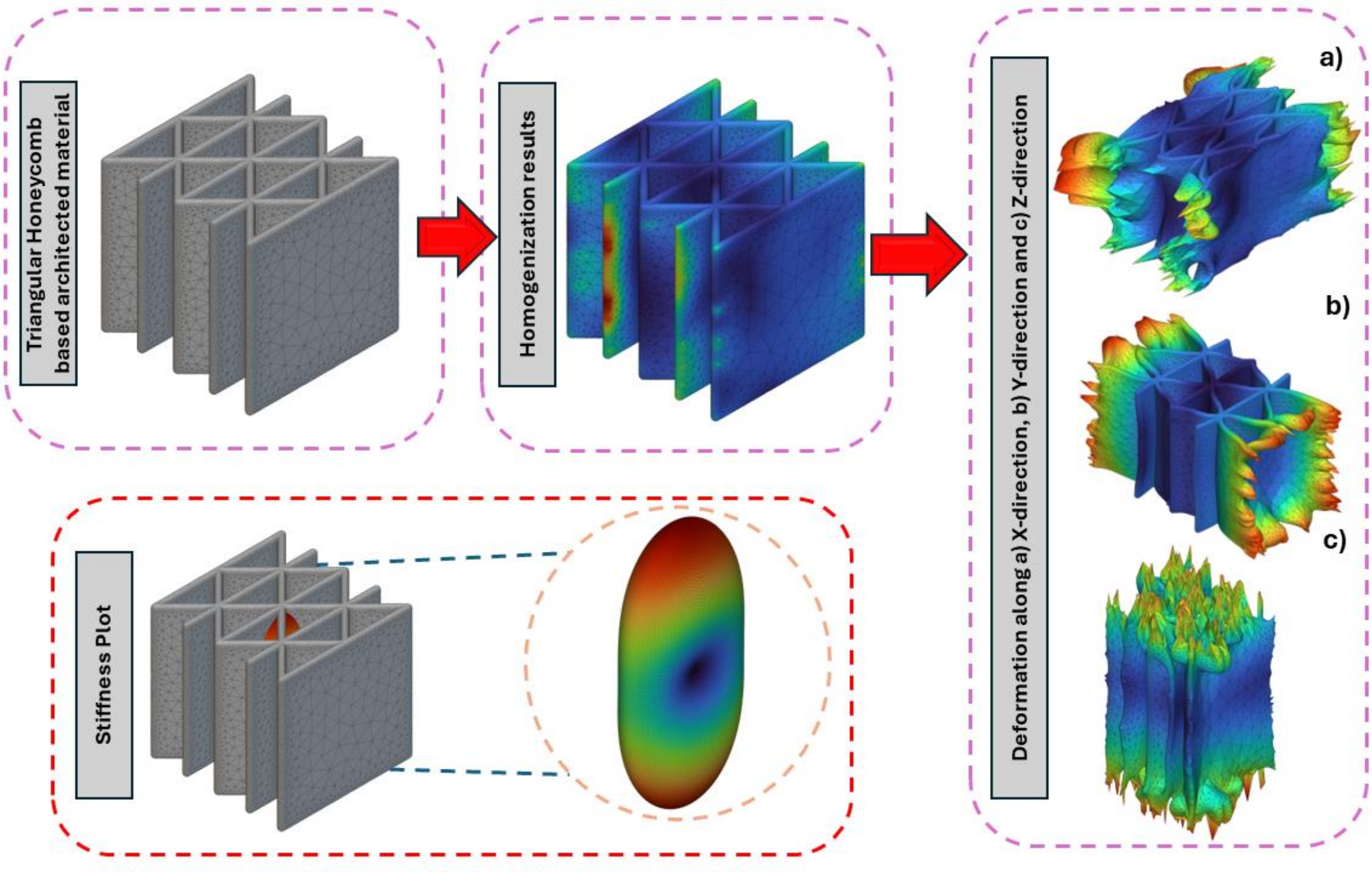
Triangular Honeycomb based architected material
Homogenization results
Deformation along a) X-direction, b) Y-direction and c) Z-direction
a)
b)
c)
Stiffness Plot

k)

**Figure 4.** Visualization of the stress and strain fields within the a) BCC, b) Diamond, c) FCC, d) FCC Foam, e) Hexagonal honeycomb, f) Iso Truss, g) Kelvin cell, h) Octet, i) Re-entrant honeycomb, j) Simple cubic and k) Traingular honeycomb architected material microstructure. This figure presents the results of the homogenization analysis, displaying the distribution of stress and strain components throughout the unit cell geometry.

### 3.2.Finding the best model

In the present work, five supervised machine learning regression-based algorithms i.e. decision trees, cat boost regressor, XG Boost regressor, extra tree regressor, and gradient boosting regressor are implemented for predicting the effective Young's modulus value.

Regression problems involve the prediction of continuous outcomes based on input features, and one sort of decision tree model used for these tasks is the DecisionTreeRegressor. max_depth and random_state are two of the hyperparameters listed in the DecisionTreeRegressor class. The max depth of the decision tree, which establishes the model's complexity, is controlled by the max depth hyperparameter. While a deeper tree can help identify more intricate patterns in the data, it can also cause overfitting, a situation in which the model performs well on training data but badly on untested data. Model complexity and generalization are balanced by restricting the decision tree's growth to three levels with max_depth=3. In order to anticipate the residuals of the predictions made by the preceding ensemble, each decision tree in the ensemble of decision trees that Cat Boost builds is trained to do so. Since the model is set up to use root mean squared error (RMSE) as the loss function, the method employs it for optimization. Moreover, symmetric tree development for increased prediction stability, ordered boosting for consistent treatment of categorical features, and early halting based on the validation set performance are all used by Cat Boost to prevent overfitting. The foundation of XG Boost is gradient boosting, a technique that builds an ensemble of decision trees one after the other, each one improving prediction accuracy by learning from the mistakes of the one before it. In order to minimize overfitting, XG Boost employs a regularized objective function to strike a compromise between the model's complexity and data fit. Moreover, it includes parallel processing, which makes training on big datasets effective. Multiple decision trees are constructed for regression tasks using the Extra Trees Regressor, an ensemble learning technique. Extra Trees grows each tree using the whole training dataset, as contrast to conventional decision tree-based methods that employ bootstrapping to produce subsets of the training data. In the process of creating trees, it adds more randomization by choosing thresholds at random for each node's data splitting. Because of this randomness, the trees may become more robust and diverse, which could improve generalization. The average of the forecasts made by each tree in the ensemble constitutes the final forecast. The number of trees (n_estimators=10) and the degree of randomness added during the tree-building process are two of the hyperparameters that the algorithm allows for adjustment. Gradient Boosting Regressor is a well-liked ensemble learning technique that builds a more reliable and accurate model for regression tasks by combining several weak learners, usually decision trees. It works by training each tree iteratively to reduce the residual errors of the prior trees in the ensemble, thereby fixing the faults made by those trees. The algorithm determines the gradient of the loss function (such as mean squared error) in relation to the predictions at each iteration, which directs the training procedure. Table 3 shows the obtained metric features for each algorithms and Figure 5-7 shows the actual value vs predicted value plots, Q-Q plots and residual plots.

Figure 6 shows the five Q-Q plots, each representing the relationship between the theoretical quantiles and the observed values of the residuals for different regression models. The Decision Tree Regression, XG Boost Regression, and Cat Boost Regression models all show a clear linear relationship between the theoretical and observed quantiles, suggesting that the residuals from these models follow a normal distribution. The Extra Tree Regression model's Q-Q plot shows a slightly curved pattern, which may indicate a slight deviation from normality in the residuals, but the overall trend is still linear, indicating that the residuals are reasonably well-behaved. The Gradient Boosting Regression model's Q-Q plot demonstrates a clear linear relationship between the theoretical and observed quantiles, suggesting that the residuals from this model follow a normal distribution. This suggests that the assumptions of the regression models are generally met, and the models can be considered reliable for making inferences or predictions.

**Table 3.** Metric features obtained for each algorithms

| **Algorithms** | **MSE** | **MAE** | **R-square value** |
|---|---|---|---|
| Decision Tree | 37.5061 | 3.1171 | 0.8333 |
| XG Boost | 2.7993 | 1.1521 | 0.9875 |
| Cat Boost | 6.7487 | 1.8070 | 0.9700 |
| Extra Tree | 21.2221 | 2.9399 | 0.9057 |
| Gradient Boosting | 27.2566 | 3.4953 | 0.8789 |

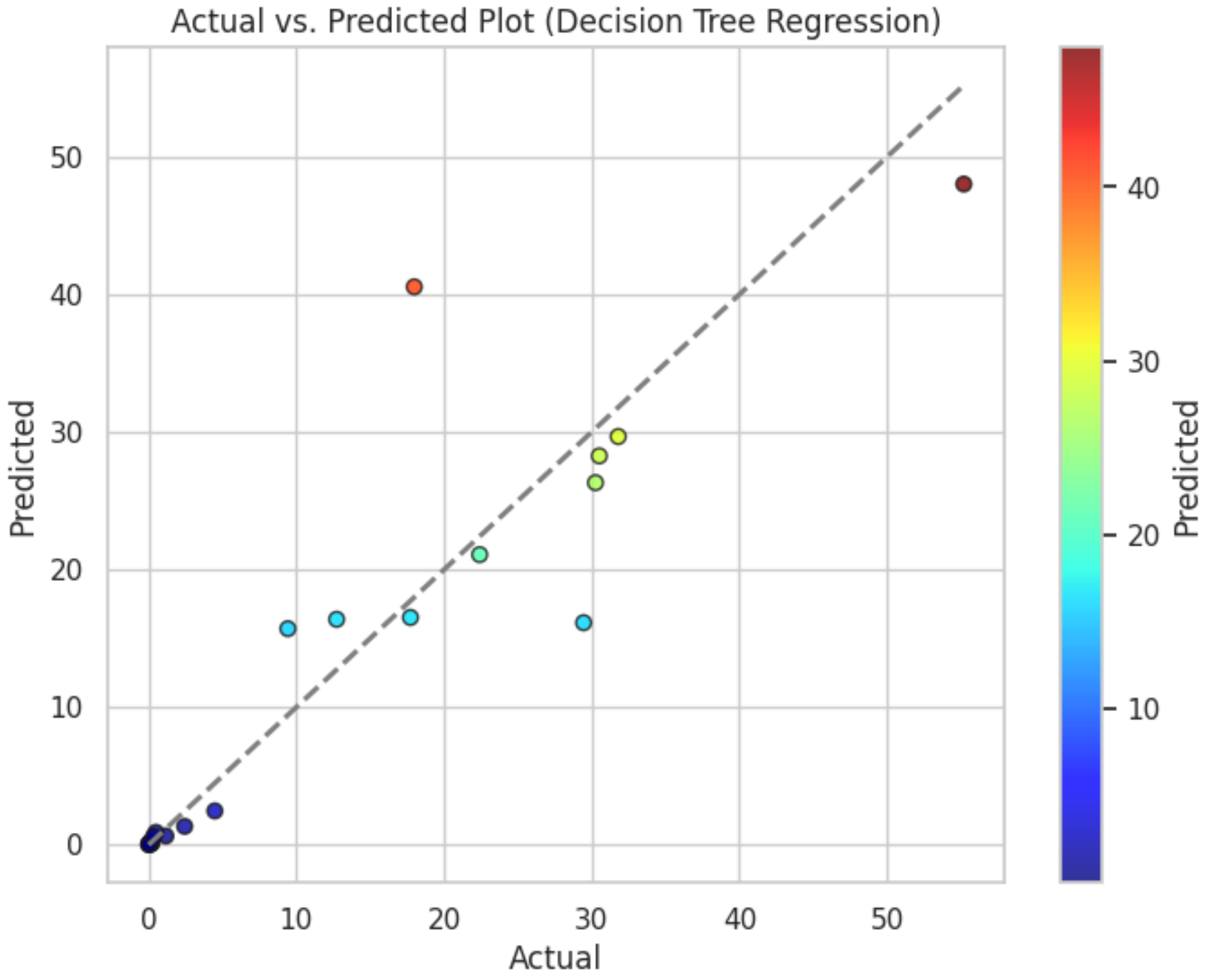

a)

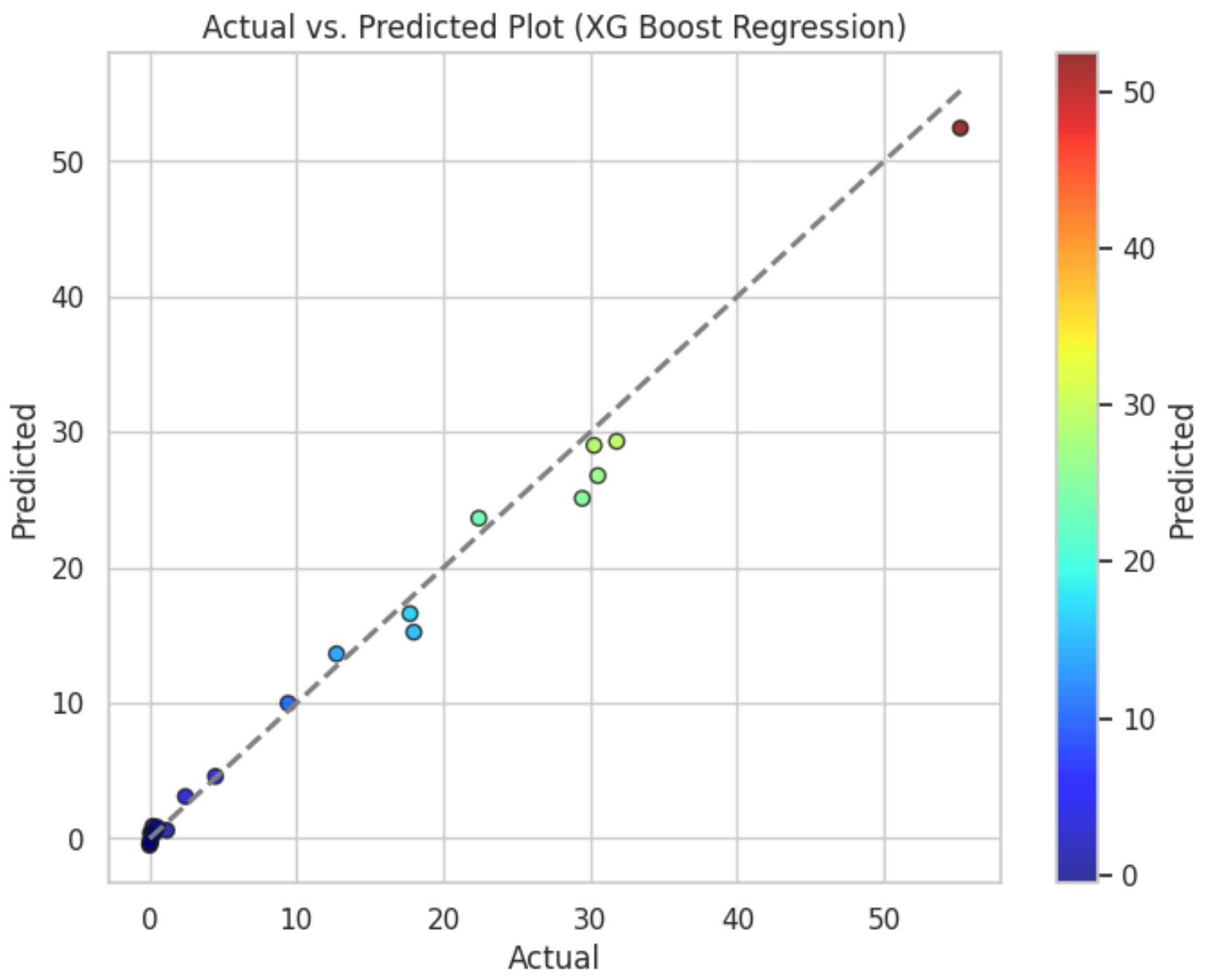

b)

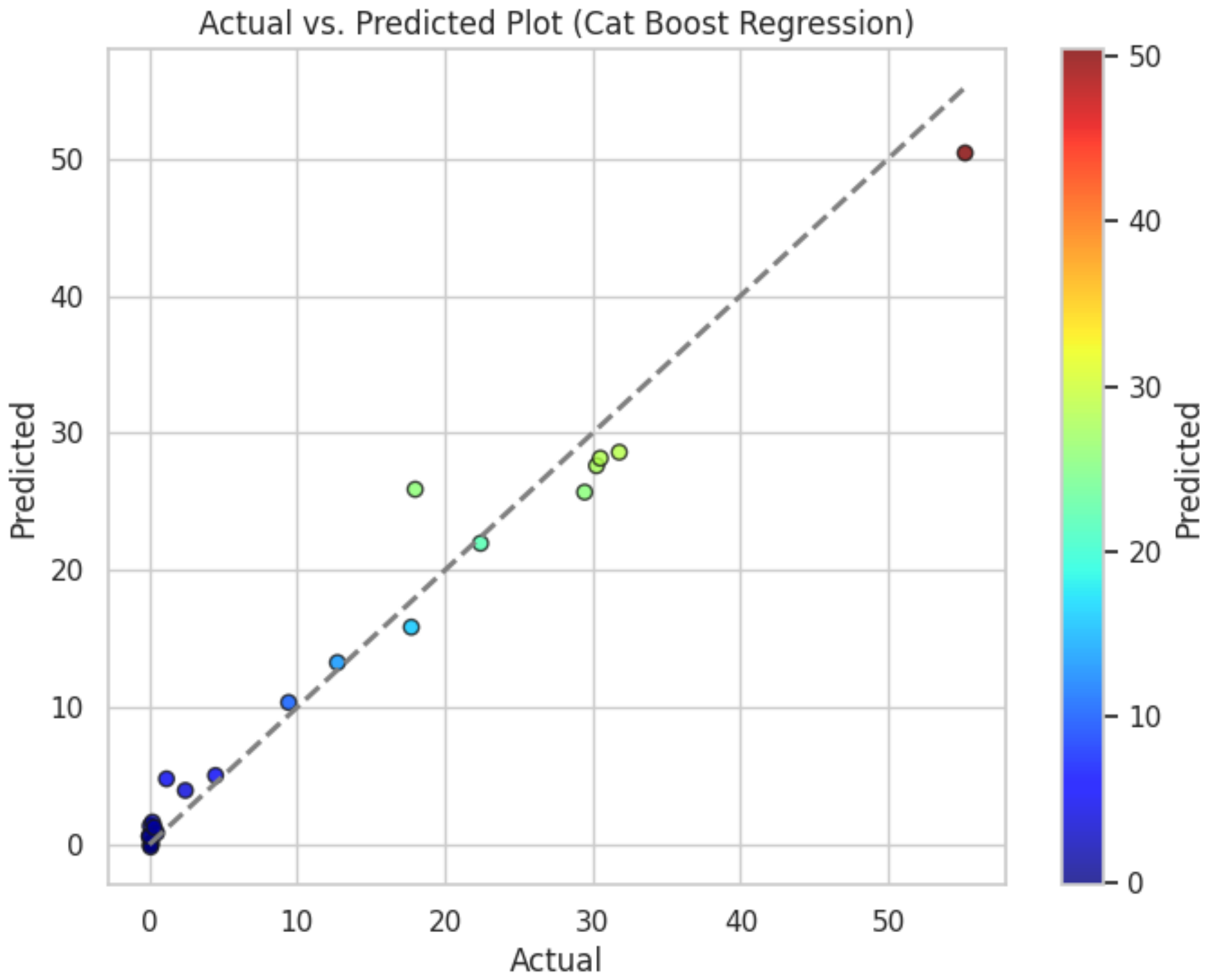

c)

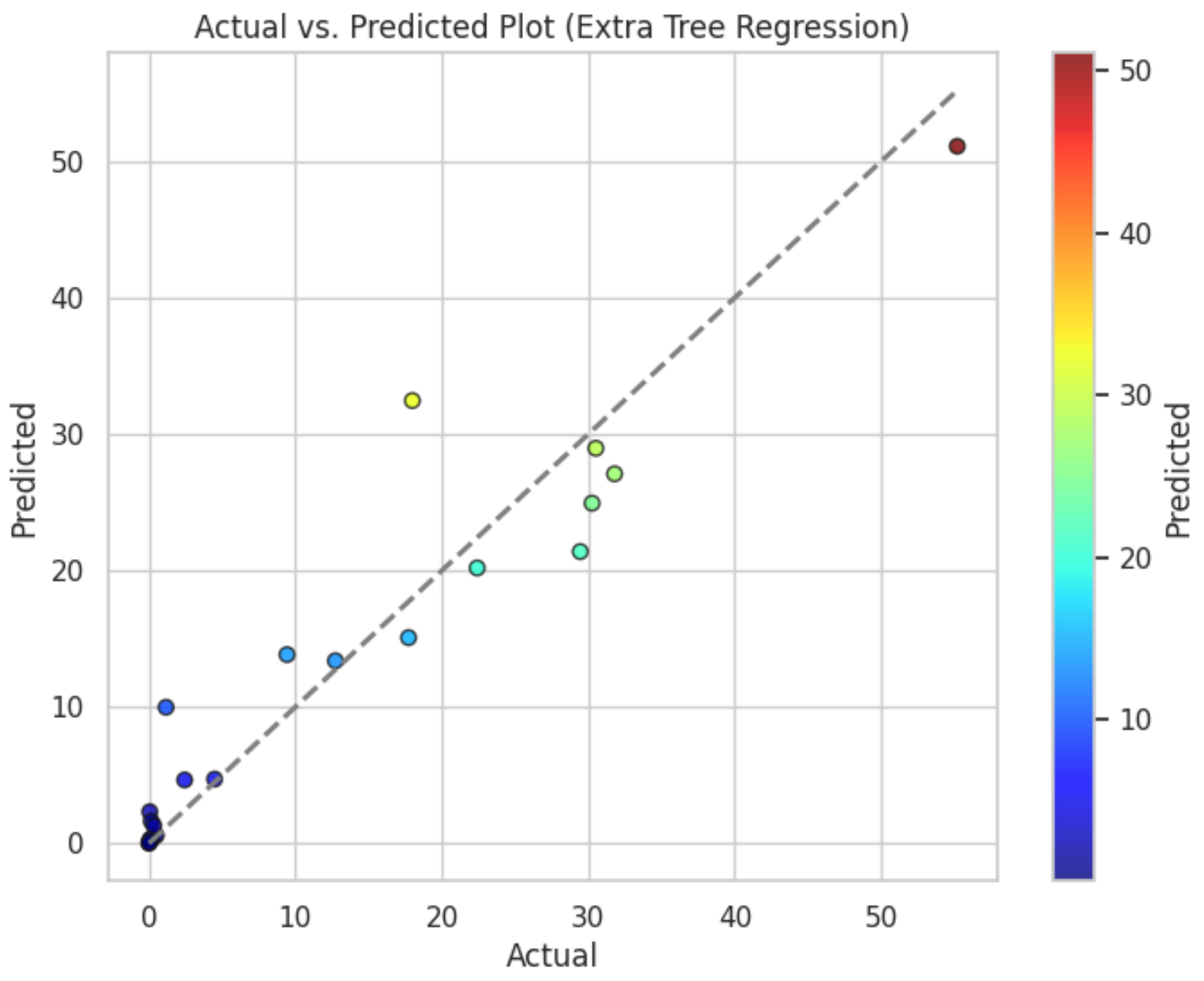

d)

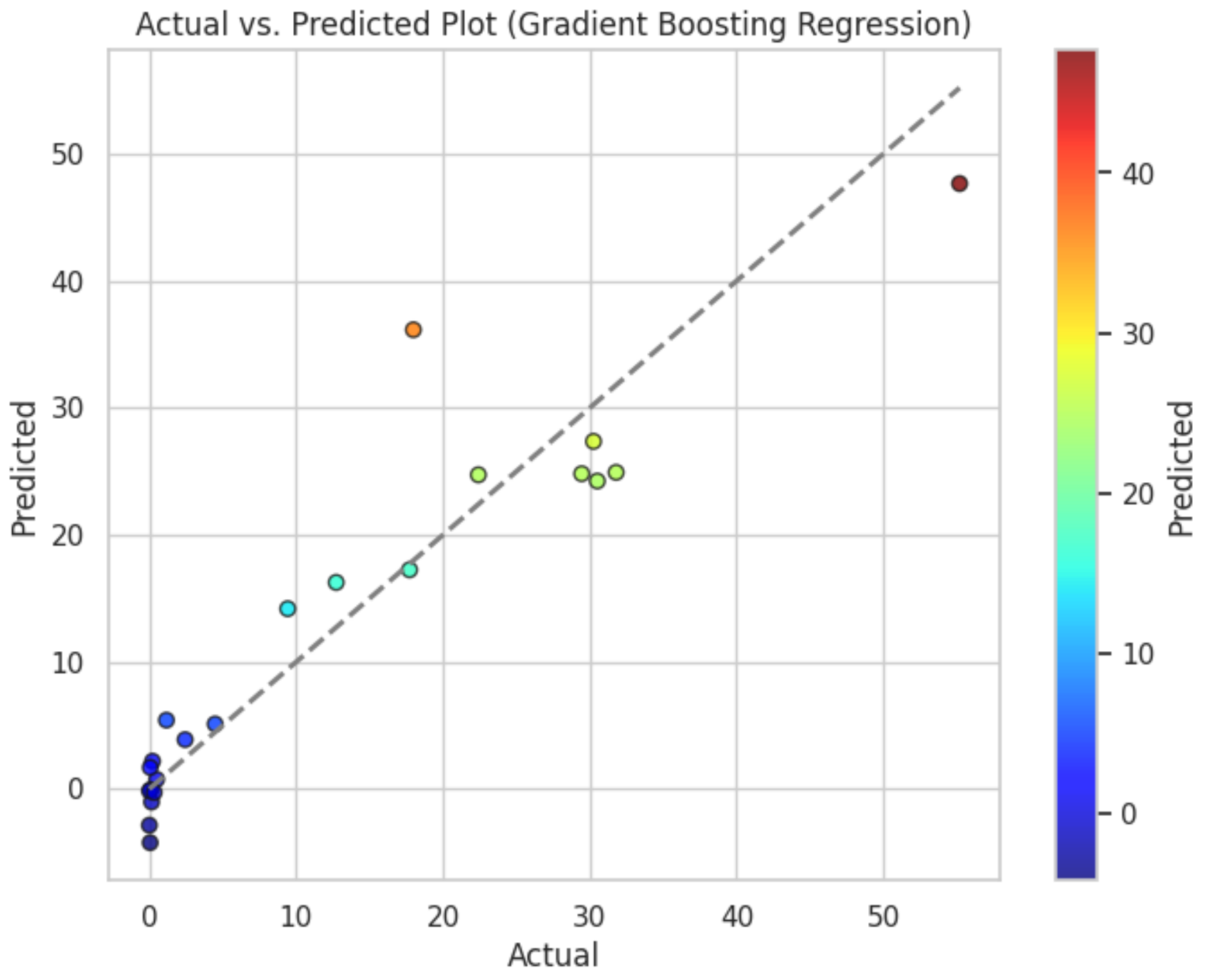

e)

**Figure 5.** Actual vs predicted effective Young's Modulus values for a) Decision trees, b) XG Boost, c) Cat boost, d) Extra tree and e) Gradient boosting regression algorithms

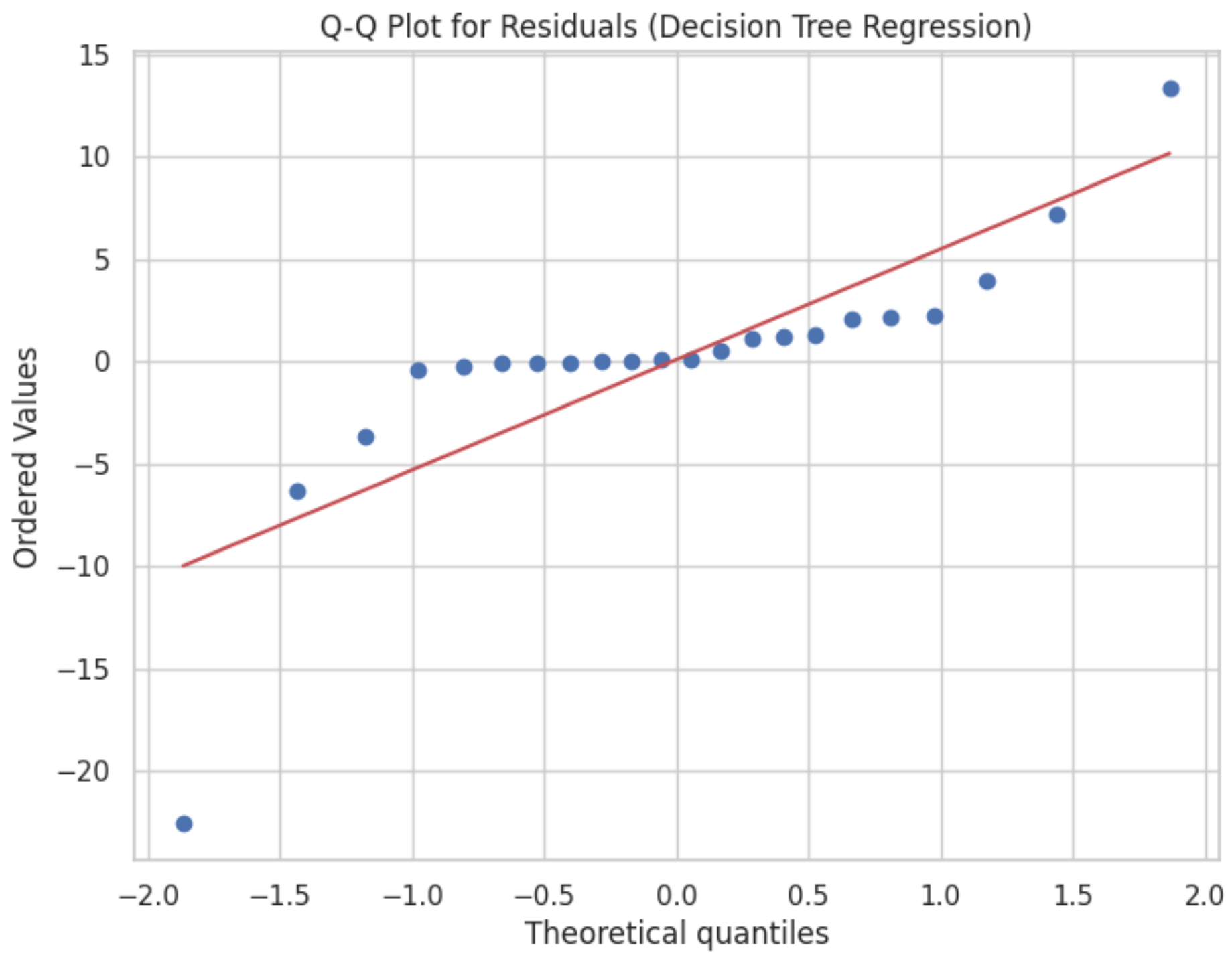

a)

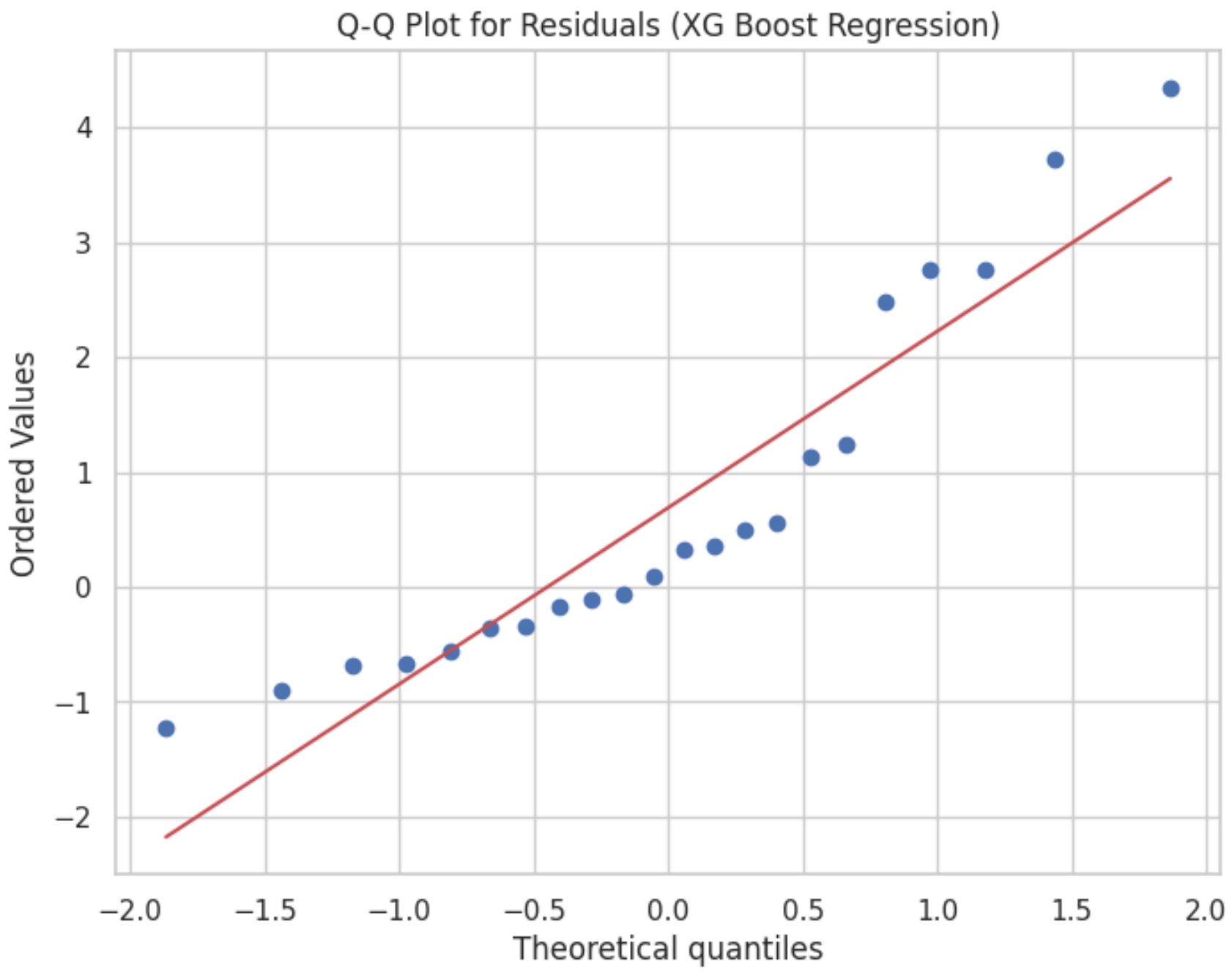

b)

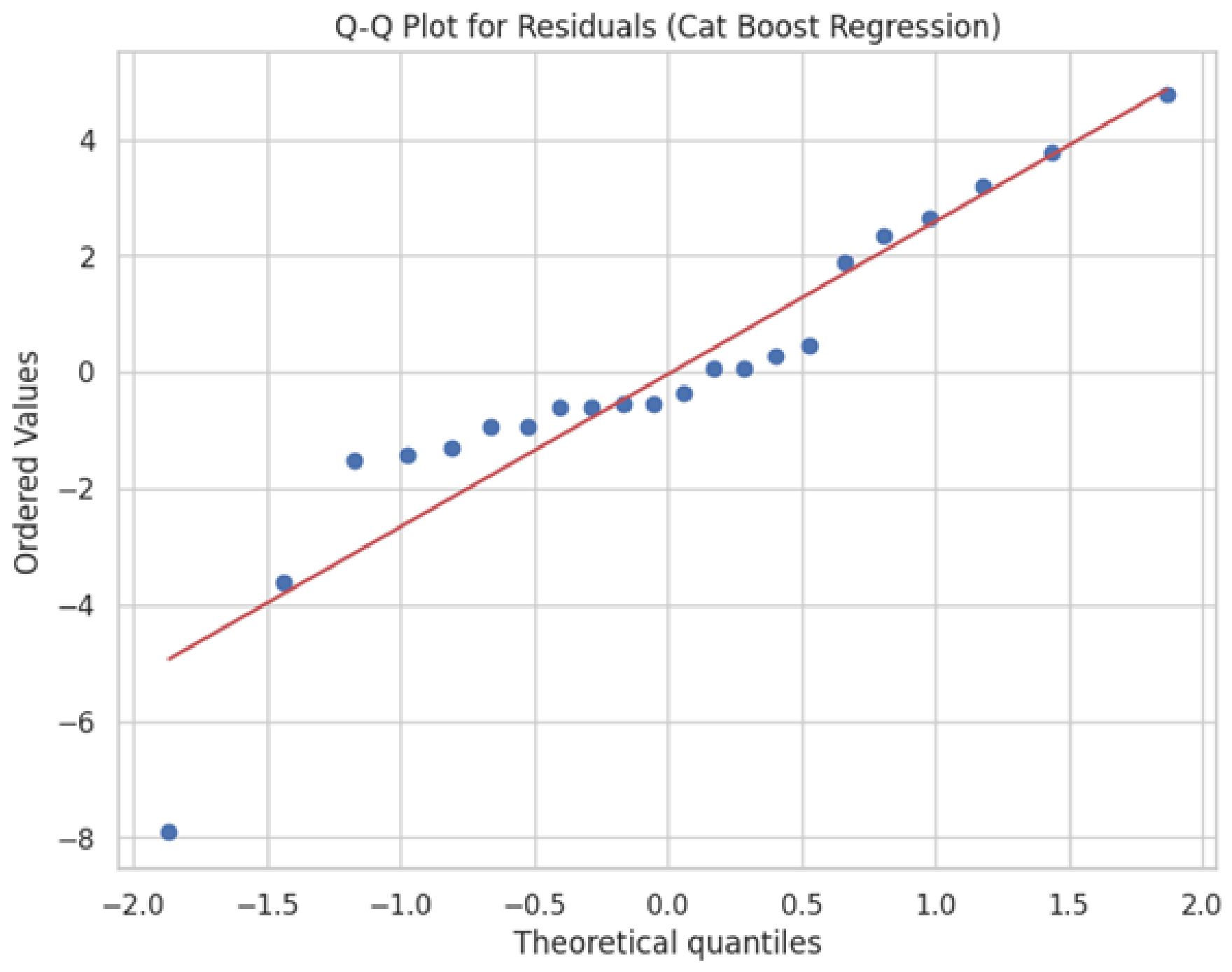

c)

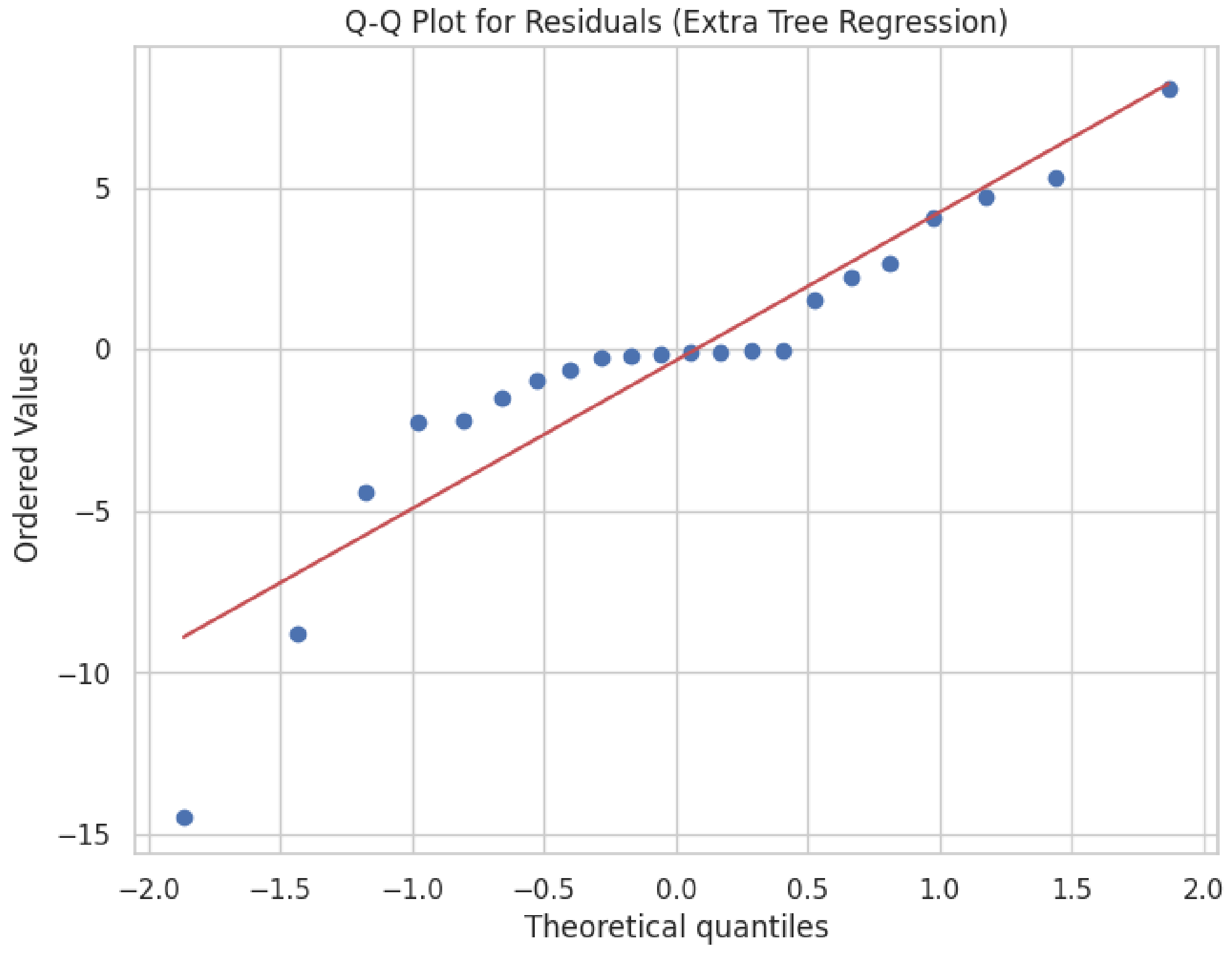

d)

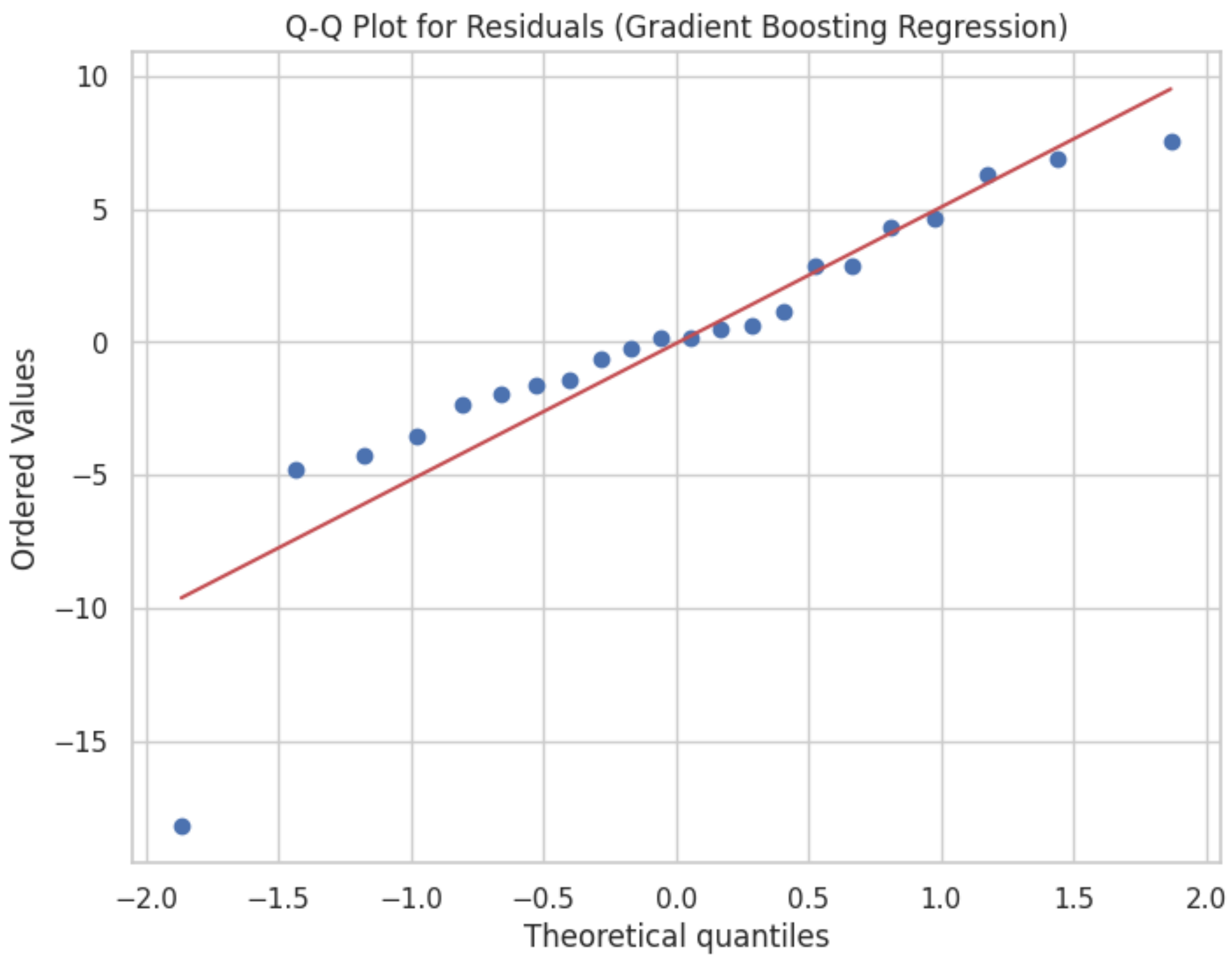

e)

**Figure 6.** Q-Q plots for a) Decision trees, b) XG Boost, c) Cat boost, d) Extra tree and e) Gradient boosting regression algorithms

The residual plot for the Decision Tree Regression model shows a relatively even distribution of residuals around the zero line, suggesting a good fit of the model to the data. The residuals appear to be randomly scattered, indicating that the model assumptions are reasonably well met. The residual plot for the XG Boost Regression model also exhibits a relatively even distribution of residuals around the zero line, with a few outliers. This suggests that the XG Boost Regression model provides a reasonably good fit to the data, though there may be some room for improvement. The residual plot for the Cat Boost Regression model displays a similar pattern to the XG Boost Regression, with an even distribution of residuals around the zero line and a few outliers. This indicates that the Cat Boost Regression model also provides a reasonably good fit to the data. The residual plot for the Extra Tree Regression model shows a more scattered pattern of residuals, with some clustering and a more pronounced curvature in the overall trend. This suggests that the Extra Tree Regression model may not fit the data as well as the other models. Finally, the residual plot for the Gradient Boosting Regression model demonstrates a relatively even distribution of residuals around the zero line, with a few outliers. This suggests that the Gradient Boosting Regression model provides a good fit to the data, similar to the Decision Tree and XG Boost Regression models.

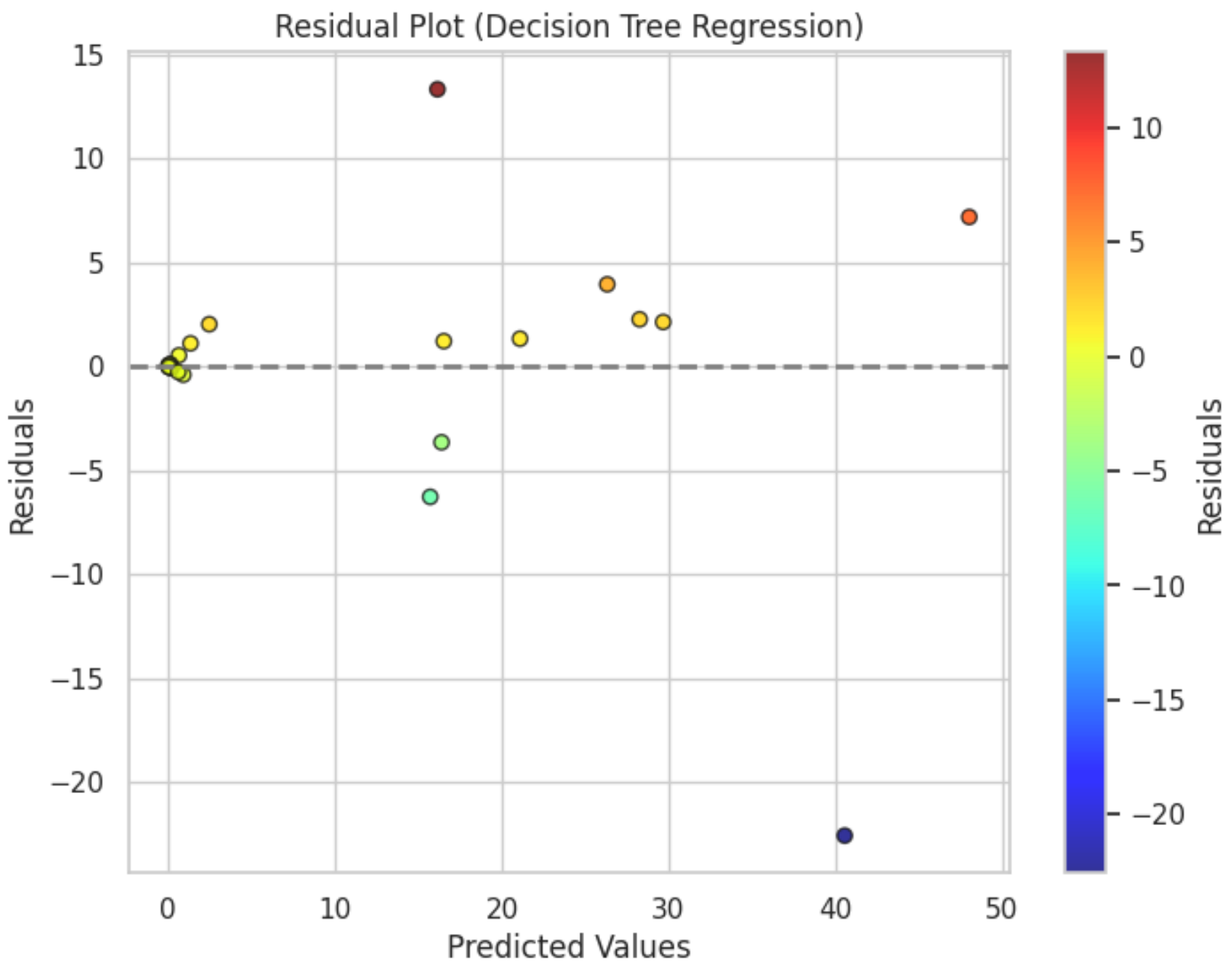

a)

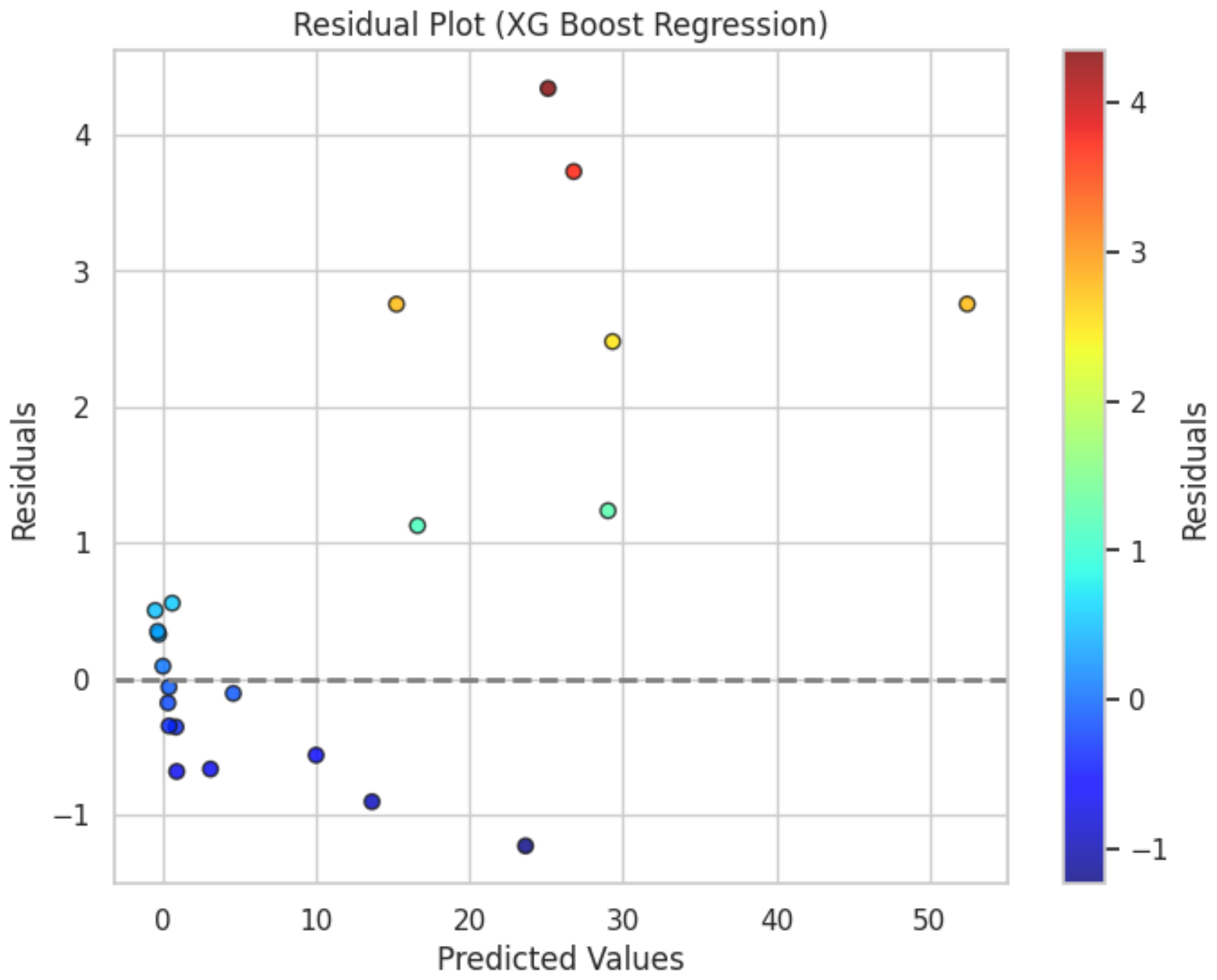

b)

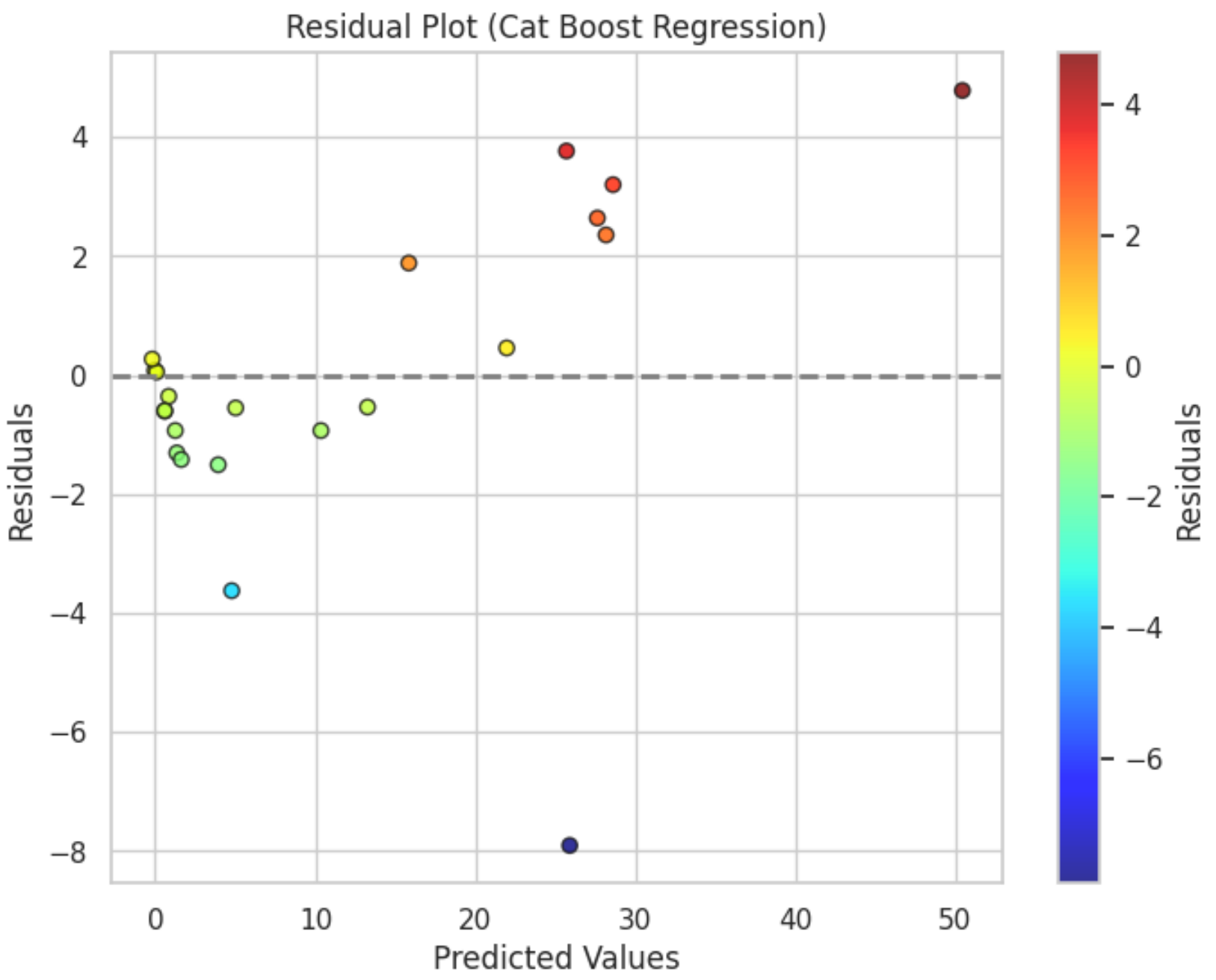

c)

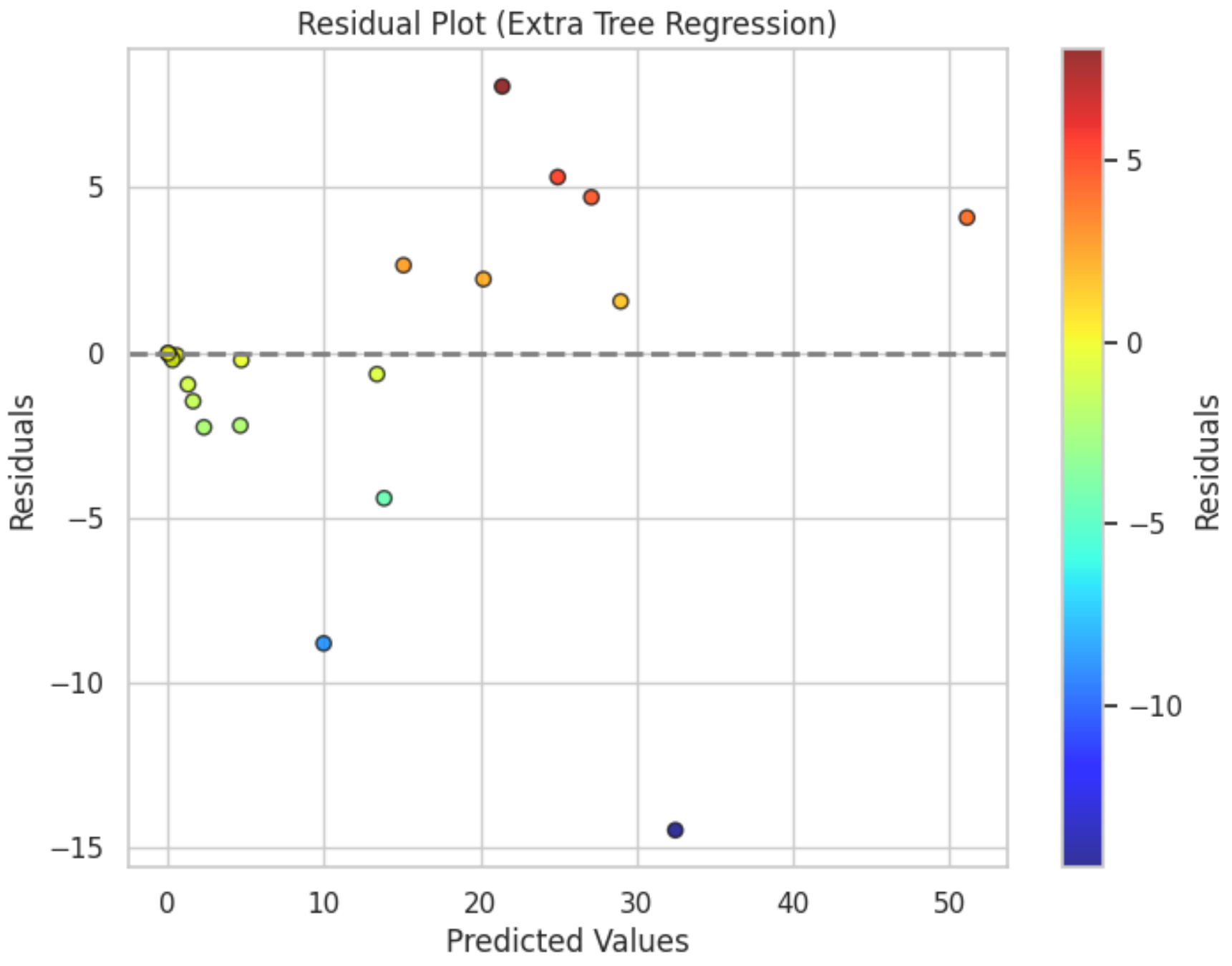

d)

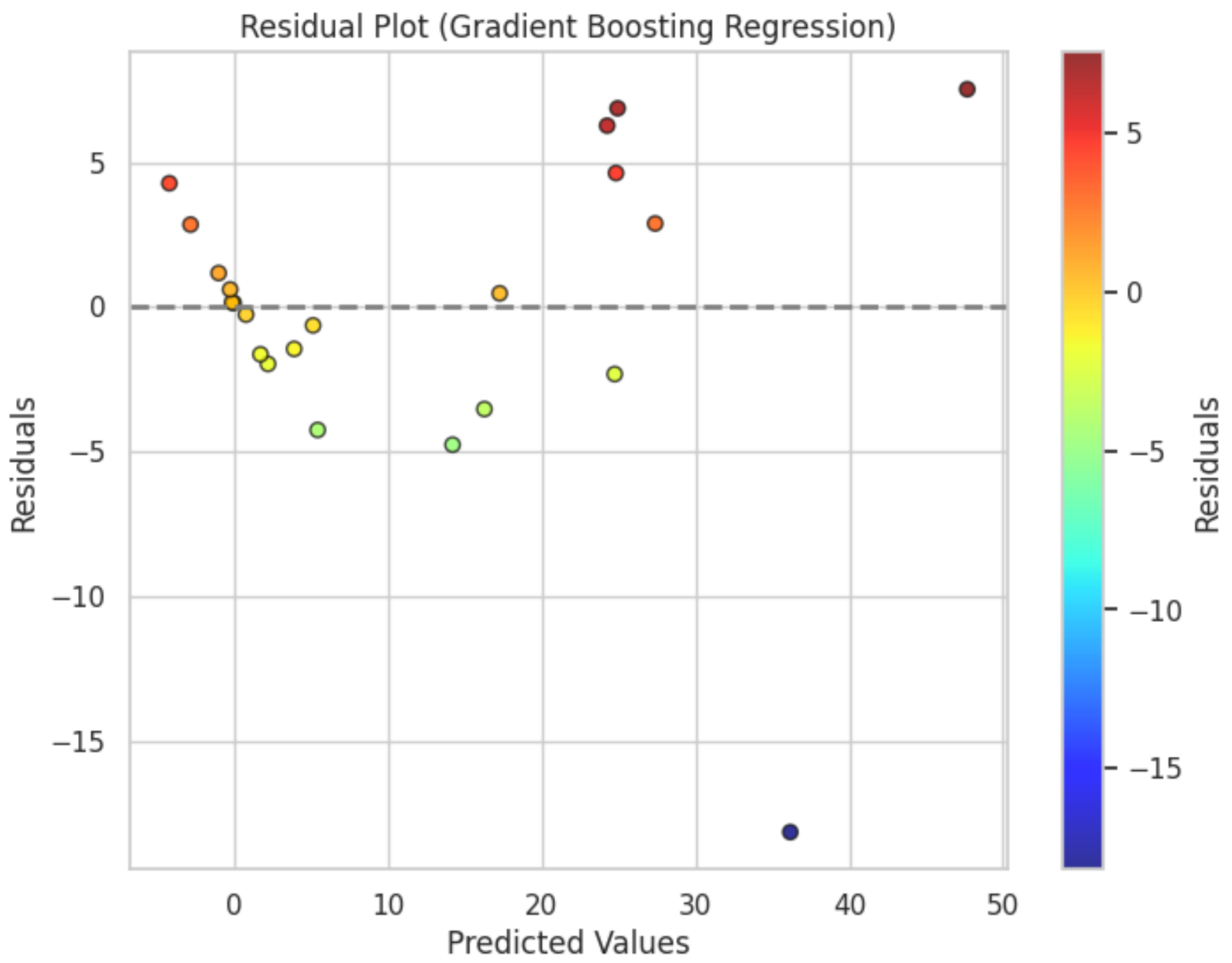

e)

**Figure 7.** Residual plots for a) Decision trees, b) XG Boost, c) Cat boost, d) Extra tree and e) Gradient boosting regression algorithms

### 3.3. Deployment of the best Machine learning model to predict the effective Young modulus

The deployment of machine learning models is the process for making models available in production environments, where they can provide predictions to the other software systems. It is the last stage in the machine learning lifecycle. There are two types of environments in the ML pipelines i.e. research environment and production environment shown in Figure 8. The investigation, testing, and development stages of a machine learning project are the main areas of concentration for the research environment. Here, researchers and data scientists develop and improve machine learning models. The final, highly-optimized machine learning models are implemented for practical application in the production environment. Its main goal is to give end users or systems scalable and dependable predictions.

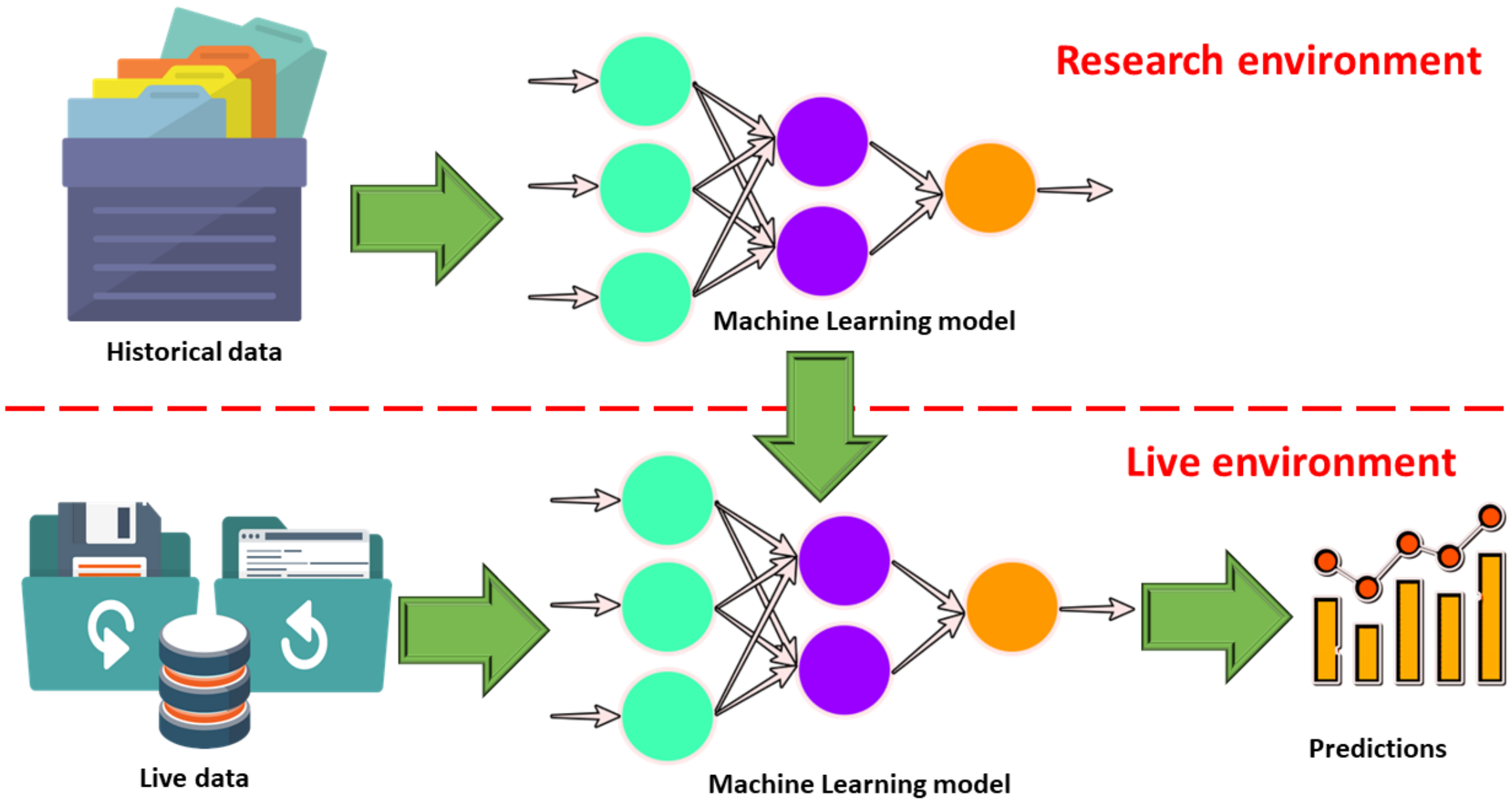

**Figure 8.** A diagram of a machine learning model deployment process, highlighting the distinction between research and production environments, and the key components required for successful integration of machine learning models in a production system.

We develop our machine learning model in a research environment which is an isolated model and here we train our machine learning model on the available historical data. If we are satisfied with the obtained results from the research environment then we are ready to migrate it to the production environment where it receive the live data and make further prediction. It should be noted that our machine learning models should be reproducible in both research and production environment. Reproducibility guarantees that when the model is trained and assessed in various settings, the same outcomes will be obtained. During the study phase, it is simple to verify experimental results thanks to this consistency. When other researchers can repeat the experiments and get comparable results, the research findings are considered validated. Transparency and trust in the dependability of the machine learning models are promoted by reproducibility in the research setting.

Machine learning in production require different multiple components like data, documentation, infrastructure, applications and configuration to work properly. The general layout and composition of a system that integrates machine learning models and algorithms is referred to as the machine learning (ML) system architecture. It includes all of the different parts, sections, and exchanges required to create, implement, and oversee machine learning systems. The performance, scalability, maintainability, and general success of an ML system are significantly influenced by its architecture. Production code is designed to be deployed for end users. The key area of machine learning system is shown in Figure 9.

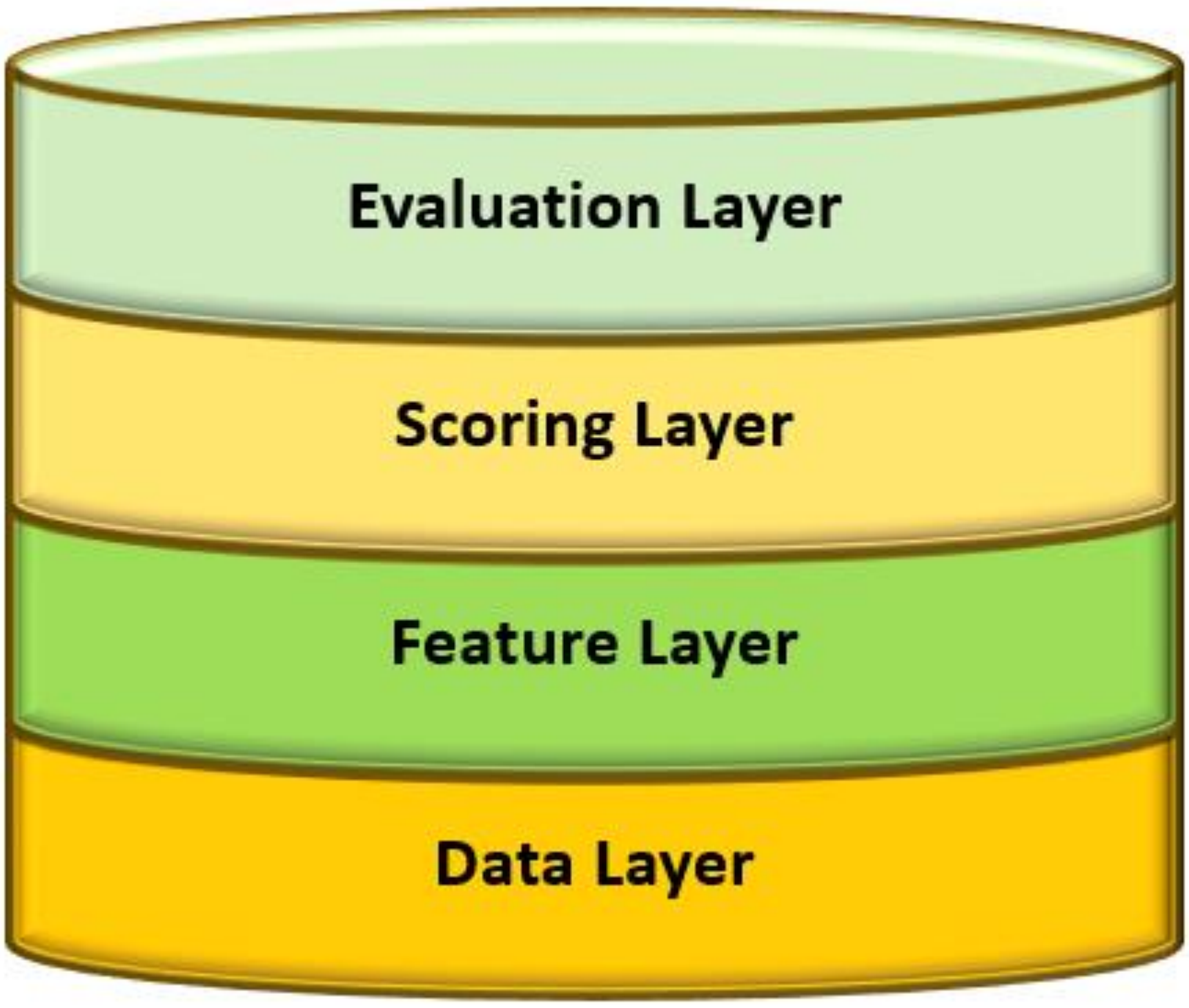

**Figure 9.** The general layout and composition of a machine learning system architecture, including the data, feature, scoring, and evaluation layers, which are essential for the deployment and operation of machine learning models in real-world applications.

The input data must be handled and managed by the data layer. Data collection, preprocessing, cleaning, and transformation are some of the tasks involved. The data layer makes sure the data is in an appropriate format so the machine learning model can process it further. Features in machine learning refer to the input variables or characteristics that are utilized in the prediction process. The tasks associated with feature engineering, which converts unprocessed data into a set of features the machine learning model can use to identify trends and anticipate outcomes, are included in the feature layer. A trained machine learning model is usually used in the scoring layer to make predictions or score new data points. To produce predictions or scores, the model must be applied to the input features. The scoring layer generates class labels or probabilities in classification problems and continuous predictions in regression problems. The machine learning model's performance is evaluated by the evaluation layer. In order to determine various performance metrics, such as accuracy, precision, recall, F1 score, or others depending on the problem's nature, it entails comparing the model's predictions to the actual outcomes (ground truth). The assessment layer aids in determining how well the model applies to fresh, untested data.

The provided web application is designed to predict the Young's Modulus of Architected Materials using machine learning models. Users can upload their datasets and receive predictions based on their input features. The application utilizes the Streamlit framework to create an interactive, data-driven web interface.

The application begins with a title and subtitle, followed by an image of lattice structures shown in Figure 10. Users can upload a CSV file containing data for analysis, which is then loaded into a Data Frame using pandas and pre-processed. This involves encoding categorical variables using Label Encoder and removing unnecessary columns. After preprocessing, the data is split into training and testing sets in 80-20 ratio. Features are standardized using Standard Scaler for consistent scaling across different features. An XGBoost regressor (XGBRegressor) is used as the machine learning model and trained on the training data. The application allows users to input their own features for prediction, either through numerical inputs or by selecting a value from a dropdown menu for categorical features. The user inputs are standardized and used to make predictions.

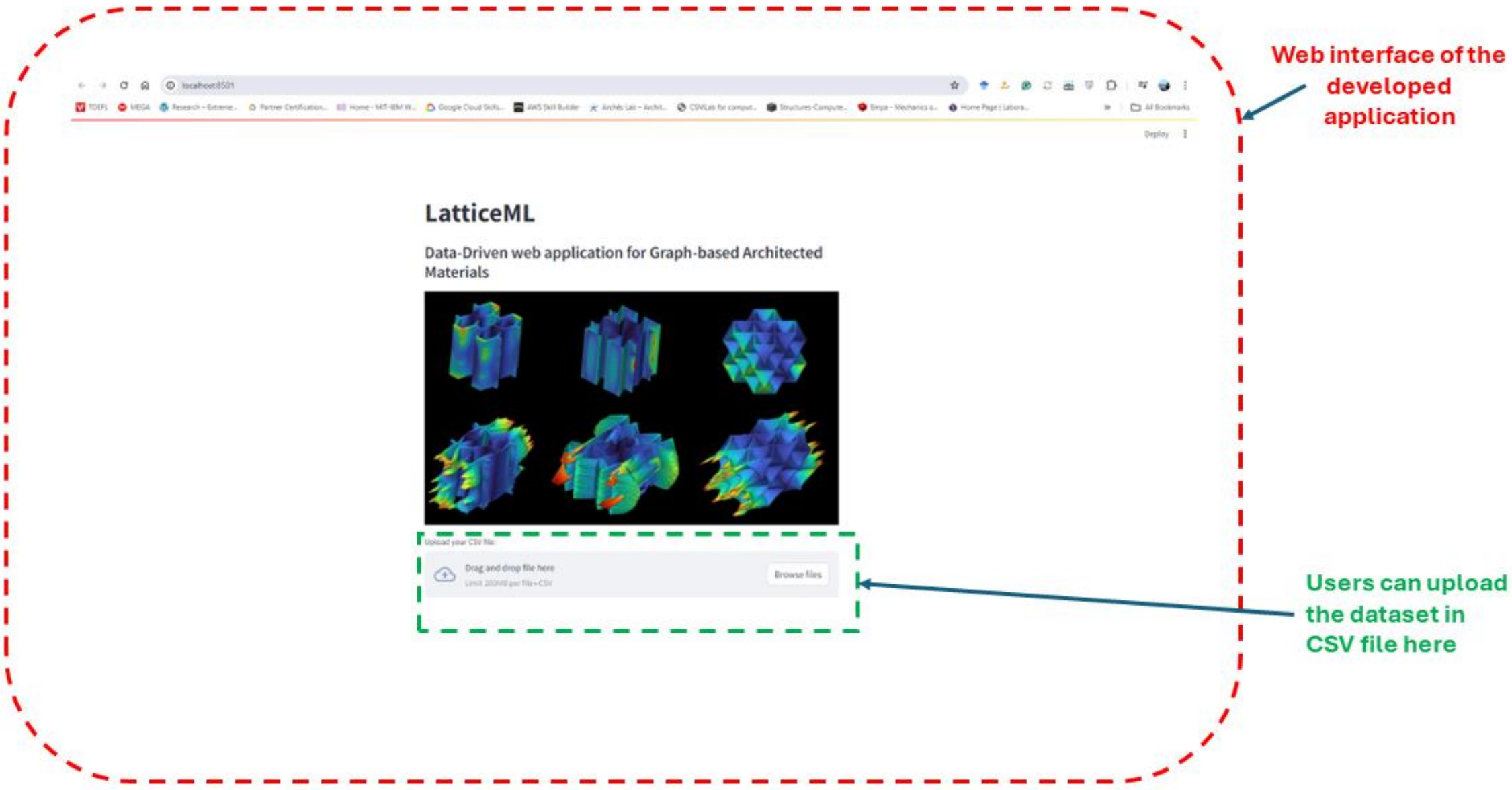

**Figure 10.** Interface of the LatticeML web application, featuring the title, subtitle, and an image of various lattice structures.

The application displays the predicted Young's Modulus of the Architected Material based on user inputs shown in Figure 11. Additionally, it calculates and presents model performance metrics such as mean squared error (MSE), mean absolute error (MAE), and R-squared ($R^2$) error using the test data. Several visualizations are included to demonstrate the model's performance and insights into the data. These include a correlation heatmap, feature importances, an actual vs. predicted plot, residual plots, and a Q-Q plot for residuals. These visualizations help in understanding the relationships between features, the importance of different features, how well the model performs in predicting actual values, and the distribution of residuals.

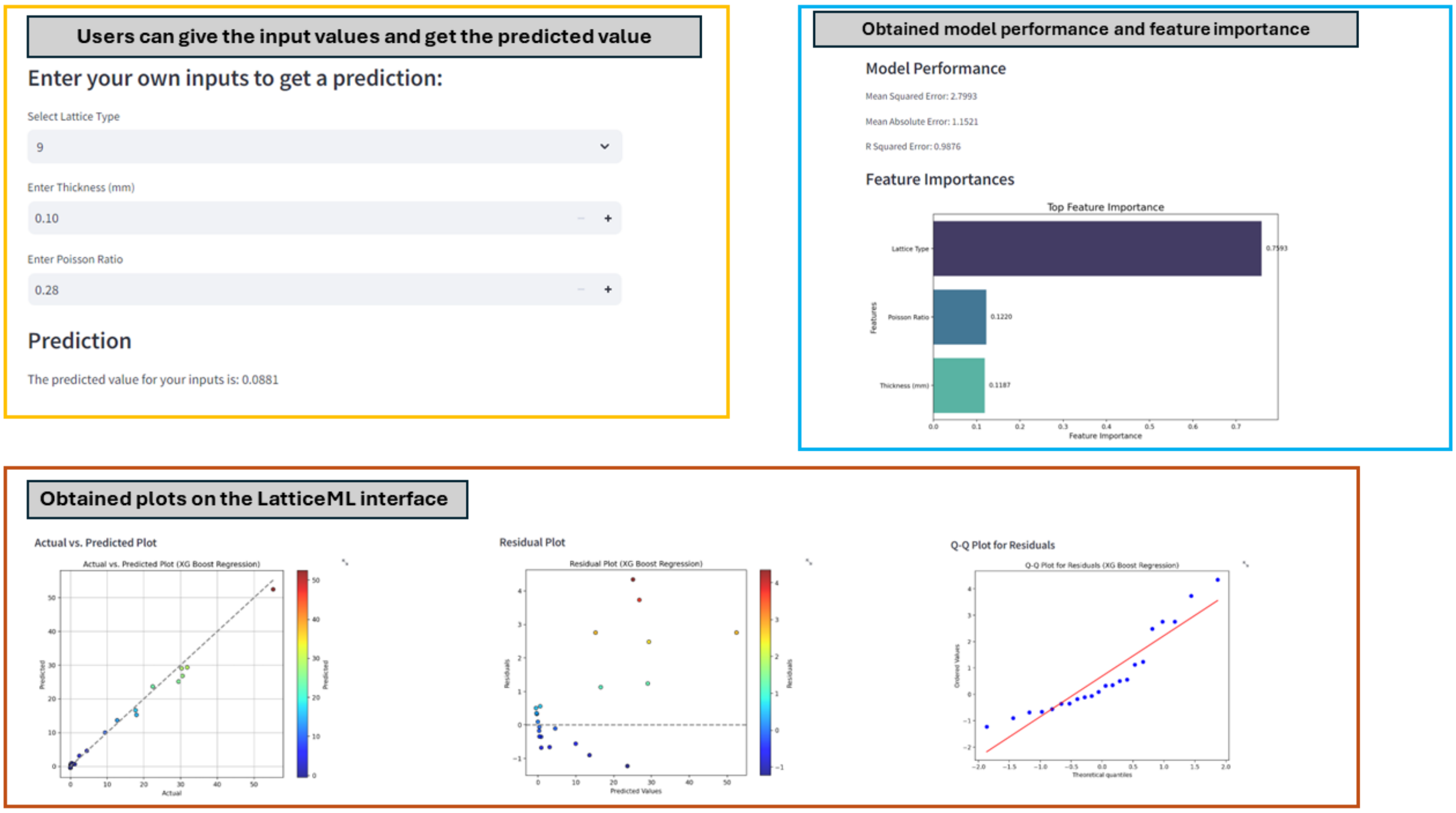

**Figure 11.** Interface of the LatticeML web application, displaying the predicted Young's Modulus value based on user inputs, as well as model performance metrics and visualizations.

## 4. Conclusion

This work has shown the degree to which the LatticeML framework predicts the effective Young's modulus of graph-based, high-temperature architected materials. This application reduces the time and resources needed for standard design and optimization procedures by combining finite element analysis with sophisticated machine learning regression models to provide fast and accurate material property predictions. With an MSE of 2.7993, MAE of 1.1521, and R-squared value of 0.9875, the results demonstrate that the XGBoost Regressor obtains the maximum prediction accuracy. One of the main advantages of the LatticeML model is its capacity to represent the complex relations between mechanical behavior, lattice geometry, and material composition. The framework's usability and accessibility are further improved by the interactive online interface, which gives engineers and designers more freedom to experiment and accelerate the creation of high-performance architected materials.

**Conflict of interest**

Author declares there is no conflict of interest.

**Funding**

This research did not receive any external funding.

**Data availability statement**

The data that support the findings of this study are available upon request from the author.